\documentclass[journal]{IEEEtran}

\ifCLASSINFOpdf
\else
\fi

\usepackage{cite}
\usepackage{times}
\usepackage{algorithmic}
\usepackage{array}
\usepackage{amssymb}
\usepackage{mdwmath}
\usepackage{mdwtab}
\usepackage{eqparbox}
\usepackage{cite}
\usepackage{url}
\usepackage{multicol,multirow}
\usepackage[cmex10]{amsmath}
\usepackage{makeidx}
\usepackage{diagbox}
\usepackage{rotating,soul}
\usepackage{wrapfig}
\usepackage{booktabs}
\usepackage[usenames,dvipsnames]{color}
\usepackage{ragged2e}
\usepackage{pifont}
\usepackage{overpic}
\makeindex

\newcommand{\zh}[1]{{\textcolor{black}{#1}}}

\usepackage{geometry}
\RequirePackage{silence}
\DeclareGraphicsExtensions{.pdf,.jpg,.png}

\usepackage[colorlinks,bookmarksnumbered,bookmarksopen,linkcolor=black,citecolor=blue,urlcolor=blue]{hyperref}

\newcommand{\figref}[1]{Fig.~\ref{#1}}

\def\ie{\emph{i.e.}}
\def\eg{\emph{e.g.}}
\def\etal{{\em et al.}}

\newcommand{\addFig}[1]{}
\newcommand{\addFigs}[1]{}

\graphicspath{{./figs/}}

\newlength\savedwidth

\begin{document}

\title{Learnable Prompting SAM-induced Knowledge Distillation for Semi-supervised Medical Image Segmentation}

\author{
Kaiwen Huang, 
Tao Zhou,
Huazhu Fu,
Yizhe Zhang, Yi Zhou, 
Chen Gong,
Dong Liang
 	\\	 	
 	
\thanks{K. Huang, T. Zhou, Y. Zhang and C. Gong are with the School of Computer Science and Engineering, Nanjing University of Science and Technology, Nanjing 210094, China. Corresponding author: \textit{Tao Zhou} (taozhou.dreams@gmail.com).}
\thanks{H. Fu is with the Institute of High-Performance Computing, Agency for Science, Technology and Research, Singapore.}
\thanks{Y. Zhou is with the School of Computer Science and Engineering, Southeast University, Nanjing 211189, China. }
\thanks{D. Liang is with the Lauterbur Research Center for Biomedical Imaging and the Research Center for Medical AI, Shenzhen Institute of Advanced Technology, Chinese Academy of Sciences, 
Shenzhen 518055, China.}
}

\markboth{Submitted to IEEE TMI,~Vol.~xx,No.~xx,~xxx.~xxxx}%
{Liu \MakeLowercase{\textit{et al.}}: Dynamic Feature Integration for Simultaneous Detection of Salient Object, Edge and Skeleton}

\maketitle


\begin{abstract}

The limited availability of labeled data has driven advancements in semi-supervised learning for medical image segmentation. Modern large-scale models tailored for general segmentation, such as the Segment Anything Model (SAM), have revealed robust generalization capabilities. However, applying these models directly to medical image segmentation still exposes performance degradation. In this paper, we propose a learnable prompting SAM-induced Knowledge distillation framework (KnowSAM) for semi-supervised medical image segmentation. Firstly, we propose a Multi-view Co-training (MC) strategy that employs two distinct sub-networks to employ a co-teaching paradigm, resulting in more robust outcomes. Secondly, we present a Learnable Prompt Strategy (LPS) to dynamically produce dense prompts and integrate an adapter to fine-tune SAM specifically for medical image segmentation tasks. Moreover, we propose SAM-induced Knowledge Distillation (SKD) to transfer useful knowledge from SAM to two sub-networks, enabling them to learn from SAM's predictions and alleviate the effects of incorrect pseudo-labels during training. Notably, the predictions generated by our subnets are used to produce mask prompts for SAM, facilitating effective inter-module information exchange. Extensive experimental results on various medical segmentation tasks demonstrate that our model outperforms the state-of-the-art semi-supervised segmentation approaches. Crucially, our SAM distillation framework can be seamlessly integrated into other semi-supervised segmentation methods to enhance performance. The code will be released upon acceptance of this manuscript at \href{https://github.com/taozh2017/KnowSAM}{https://github.com/taozh2017/KnowSAM}.

\end{abstract}


\IEEEpeerreviewmaketitle

\section{Introduction}

Medical image segmentation can assist doctors in observing various organs, diseased tissues, or areas of interest in patients more clearly, thus improving the accuracy of disease diagnosis~\cite{zhang2017deep,zhou2024uncertainty}. Moreover, qualitative and quantitative analysis of medical images can reveal new disease characteristics and treatment methods, enabling the evaluation of treatment effectiveness and timely adjustments to treatment plans~\cite{tajbakhsh2020embracing}. Annotating large-scale pixel-level medical data for fully-supervised training is a costly and time-consuming task. This constraint has sparked a surge of interest in semi-supervised learning (SSL) approaches~\cite{bai2023bidirectional, gao2023correlation}, which leverage a vast amount of unlabeled data to enhance model performance.

\begin{figure}[!t]
	\centering
	\begin{overpic}[width=0.44\textwidth]{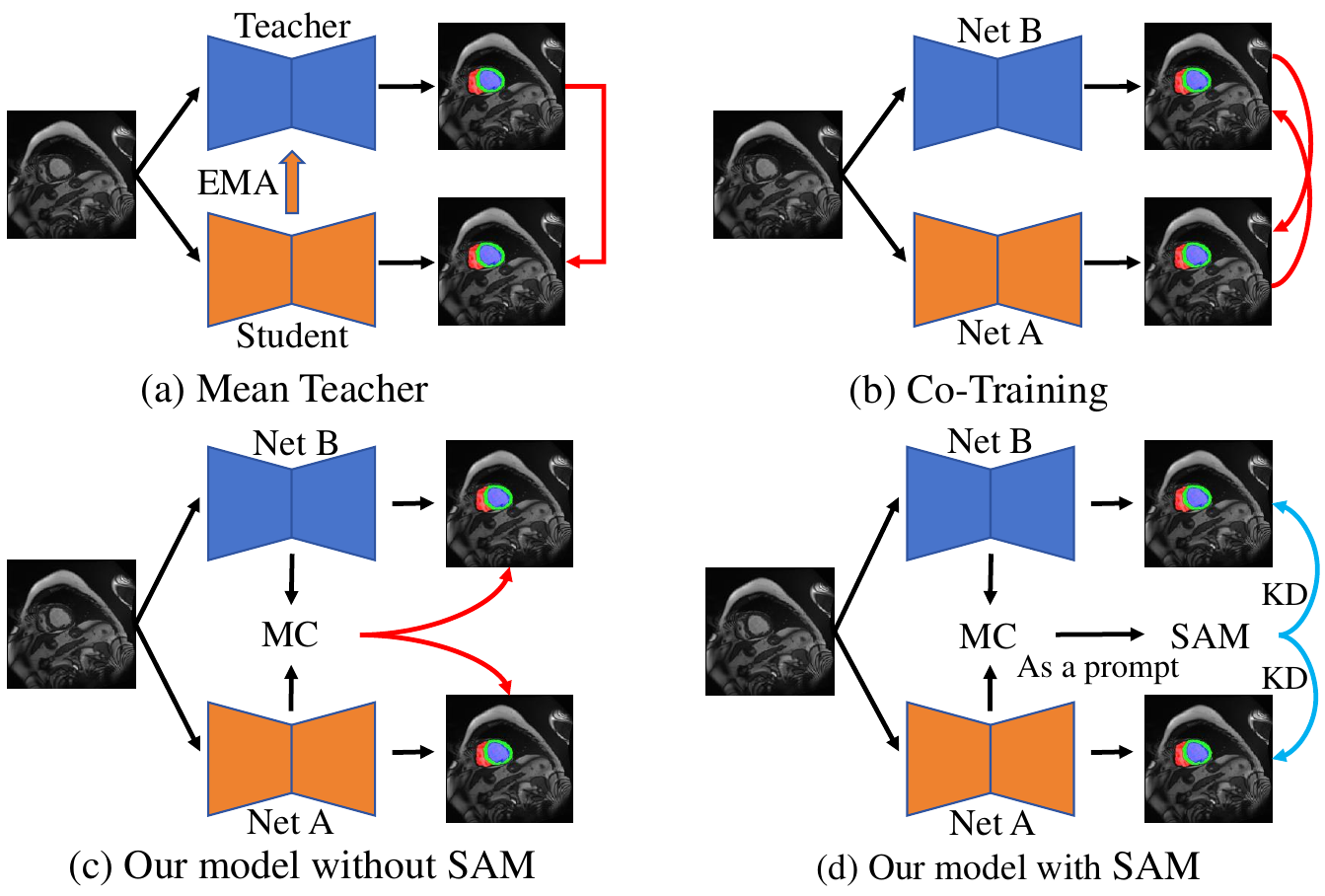}
    \end{overpic}\vspace{-0.25cm}
	\caption{{A comparison of different semi-supervised frameworks: (a) MT framework with a student model and a teacher model, (b) co-training framework with two subnets, (c) our proposed semi-supervised framework without SAM, which enhances the dual-stream network with Multi-view Co-training (MC), and (d) our model with Knowledge Distillation (KD) from SAM.
 }}
    \label{fig1}\vspace{-0.25cm}
\end{figure}

Numerous efforts have been dedicated to semi-supervised medical image segmentation~\cite{luo2021semi,luo2022semi,wu2021semi,wang2023mcf,bai2023bidirectional}. Consistency regularization has emerged as a popular technique in SSL. One notable method, mean teacher (MT)~\cite{tarvainen2017mean}, involves a student network that updates parameters through gradient propagation and a teacher network that updates parameters via exponential moving average (EMA), as shown in \figref{fig1}(a). Despite the promising results achieved by MT-based methods~\cite{xu2023ambiguity,yu2019uncertainty,wang2022semi}, they often overlook the impact of cognitive biases within the model. Subsequently, as illustrated in \figref{fig1}(b), various methods employ a co-training strategy that utilizes two subnets and explores the fusion of their outputs. Nevertheless, due to the absence of ground truth label supervision for unlabeled data, it is crucial to mitigate the impact of disagreement and the generation of low-quality pseudo-labels that may arise during co-training~\cite{shen2023co,wang2023mcf}. Moreover, other semi-supervised methods acquire prior knowledge through self-supervised learning strategies. For instance, Hu~\etal~\cite{hu2021semi} enhanced the model's understanding of latent image representations through contrastive learning during the pre-training phase. He~\etal~\cite{he2020dense} developed an autoencoder to extract deep prior anatomy (DPA) features and incorporated them into the model to enhance segmentation performance. Despite the progress that has been achieved, these methods still face numerous challenges, particularly in further enhancing the model's generalization capability.

Recently, foundational models like the Segment Anything Model (SAM)~\cite{kirillov2023segment} have demonstrated promising progress in segmentation tasks. However, SAM cannot be directly applied to all segmentation scenarios, especially in medical image segmentation tasks~\cite{chen2023sam}. To address this issue, several studies~\cite{chen2023sam,wu2023medical,li2024concatenate,xu2024esp} integrate an adapter into SAM and subsequently fine-tune SAM on the specific dataset for downstream tasks. 
Furthermore, some studies~\cite{zhang2023semisam,zhang2023segment} have incorporated SAM to enhance the accuracy of pseudo-labels for unlabeled data, thereby improving model performance. These methods supply SAM with precise prediction points or boxes as prompts to generate pseudo-labels. 
However, a limitation of this strategy is that providing incorrect prompts can directly lead to a substantial decrease in SAM's effectiveness. While numerous studies have integrated SAM into medical segmentation tasks, the optimal utilization of SAM's generalization capability to enhance semi-supervised medical image segmentation remains an area deserving of further investigation.

To this end, as depicted in \figref{fig1}(d), we propose a learnable prompting SAM-induced knowledge distillation framework (termed KnowSAM) for semi-supervised medical image segmentation. KnowSAM leverages the generalization ability of SAM through distillation learning to enhance segmentation performance. Specifically, we propose a Multi-view Co-training (MC) strategy, where two subnets engage in a co-teaching paradigm and mutually correct each other. Additionally, the MC incorporates a Hybrid Aggregation Module (HAM) to adaptively merge prediction maps with entropy maps and dissimilarity maps, reducing discordant predictions and uncertainties between the two subnets. Furthermore, we propose a Learnable Prompt Strategy (LPS) to extract dense prompts from a designed lightweight network and integrate them into SAM's decoder. Concurrently, we introduce an adapter module in SAM's image encoder and decoder layers for fine-tuning, enhancing SAM's performance on medical segmentation tasks. Additionally, we present a simple yet effective Uncertainty-Guided Data Augmentation (UGDA) method to boost the generalization of our model by increasing diversity in unlabeled data. Lastly, we present a SAM-induced Knowledge Distillation (SKD) strategy to enable the two sub-networks to learn valuable information from SAM. 

Our main contributions are summarized as follows:
\begin{itemize}
 
\item We propose a novel KnowSAM framework to harness the generalization capabilities of the foundation model through distillation learning to improve semi-supervised medical image segmentation performance. Importantly, our SAM distillation framework can be seamlessly applied to other semi-supervised segmentation methods.


\item 
We present a multi-view co-training component to integrate multi-view maps, including the subnets' outputs, entropy uncertainty maps, and spatial information from regions where discrepancies exist in the subnets' predictions. This approach gradually reduces cognitive differences between the two subnets, leading to improved segmentation accuracy.

\item 
A learnable prompt strategy is proposed to dynamically provide effective feature prompts. This strategy prevents performance degradation of SAM caused by incorrect hard coordinates, thus improving SAM's capability in medical segmentation tasks.

\item \zh{Extensive experimental results on a diversity of medical segmentation tasks demonstrate that our model outperforms the other state-of-the-art semi-supervised methods}. 




\end{itemize}

\section{Related work}\label{sec:Related works} 



\subsection{Semi-supervised Medical Image Segmentation}


Given its potential in real-world medical applications, semi-supervised segmentation has gained significant interest from researchers. Pseudo-labeling and regularization are two common methods in semi-supervised medical image segmentation, and we discuss these related studies below. 

\subsubsection{Pseudo-labeling Methods}

Pseudo-labeling methods involve generating pseudo-annotations for unlabeled data and then using them as labels for model retraining. The main challenges of this approach include generating high-quality pseudo-labels and handling the noise interference introduced by these pseudo-labels. Zeng \etal~\cite{zeng2023ss} designed a novel double-threshold pseudo-labeling method for dual-task image segmentation and classification to enhance the joint model's utilization of unlabeled data. Yao \etal~\cite{yao2022enhancing} proposed a confidence-aware cross pseudo-supervision algorithm, which enhances the quality of pseudo labels for unlabeled images by utilizing Fourier transformations. 
Moreover, strategies such as uncertainty guidance and confidence awareness can improve the quality of generated pseudo-labels, with higher confidence typically leading to more effective pseudo-labels~\cite{jiao2023learning,wang2022ssa}. 

\begin{figure*}[!t]
	\centering
	\begin{overpic}[width=0.97\textwidth]{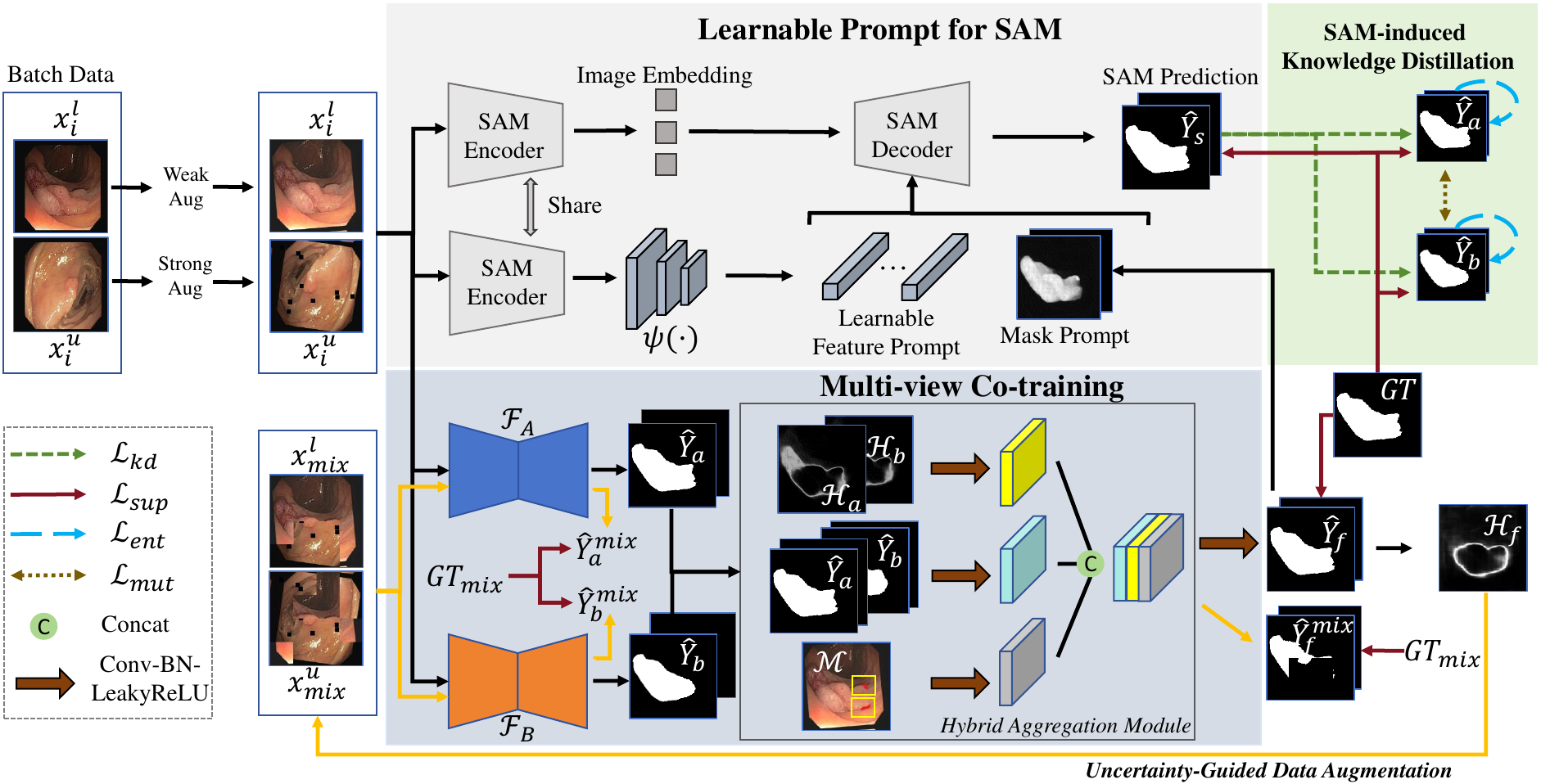}
    \end{overpic}\vspace{-0.05cm}
	\caption{{
    \zh{Overview of our KnowSAM framework. An input image is initially processed by two distinct subnets, $\mathcal{F}_A$ and $\mathcal{F}_B$ to obtain $\hat{Y}_a$ and $\hat{Y}_b$, which are then fed into a hybrid aggregation module to produce the composite map $\hat{Y}_f$. Concurrently, the input image is processed by the SAM encoder to extract feature embeddings, which are refined by $\psi(\cdot)$ to produce the learnable feature prompt. The two types of prompts ($\hat{Y}_f$ and the learnable feature prompt) are provided to the SAM decoder to produce predictions, which serve as the basis for knowledge distillation. Furthermore, we leverage an uncertainty-guided data augmentation approach to generate new training samples for enhancing robustness. 
    }.
 }}\vspace{-0.25cm}
    \label{Model}
\end{figure*}


\subsubsection{Regularization-based Methods}

Consistency regularization is a common strategy within semi-supervised methods, aiming to maintain prediction invariance under different perturbations. For example, Huang~\etal~\cite{huang2022semi} used pixel-level predictions between unlabeled samples and their perturbed counterparts for consistency regularization, enhancing the model's generalization by extracting valuable information from unlabeled data. Zheng~\etal~\cite{zheng2022double} enhanced the robustness of the network through consistency regularization by introducing random noise into the parameters of the teacher model to perturb the feature layers. Yang~\etal~\cite{yang2023revisiting} used an auxiliary feature perturbation stream to expand the perturbation space and proposed a dual-stream perturbation technique to allow two strong augmented views to be guided by a common weak view. Additionally, several semi-supervised methods~\etal~\cite{chen2021mtans,peiris2023uncertainty,wu2021collaborative} used adversarial learning to encourage the prediction distribution of unlabeled images to be as similar as possible to that of labeled data. For example, Chen~\etal~\cite{chen2021mtans} leveraged adversarial learning, combining a discriminator network and multi-scale feature consistency loss, to improve segmentation accuracy. Wu \etal~\cite{wu2021collaborative} employed adversarial learning with a focused and dispersive representation model, where the confidence maps generated by the discriminator demonstrated the effectiveness of all unlabeled data.

\subsection{Knowledge Distillation}

KD technology has achieved significant results in model compression and transfer learning, yet there has been relatively less exploration in the SSL domain. KD guides the training of student models using soft labels generated by teacher models, which typically provide more information than hard labels.
For example, Xie \etal~\cite{xie2023deep} proposed deep mutual distillation (DMD), a high-entropy online mutual distillation process that enhances semi-supervised medical image segmentation by providing more informative training than low-entropy methods, particularly in ambiguous regions. 
You \etal~\cite{you2022simcvd} introduced SimCVD, a contrastive distillation framework that improves voxel representation learning and segmentation accuracy by using unsupervised training and structural distillation of pair-wise similarities, achieving promising performance with less labeled data. 
To tackle the multi-class label imbalance challenge, You~\etal~\cite{you2023bootstrapping} proposed an anatomical-aware contrastive distillation framework for semi-supervised medical image segmentation, which leverages iterative contrastive distillation, soft negative labeling, and anatomical contrast sampling to boost segmentation accuracy.
Additionally, Shu~\etal~\cite{shu2022cross} employed a cross-mix teaching paradigm and a lightweight transductive monitor for active knowledge extraction. This serves as a bridge between the teacher and student models, improving performance in scenarios with limited supervision.

\section{METHODOLOGY}\label{Ourmethod}

\textbf{Overview}: In the semi-supervised setting, we have a labeled dataset $\mathcal{D}^L=\{(x^l_i,y^l_i)\}^{N_l}_{i=1}$ and an unlabeled dataset $\mathcal{D}^U=\{x^u_i\}^{N_l+N_u}_{i=N_l+1}$, with $N_l\ll{N_u}$. $x_i \in \mathbb{R}^{H \times W}$ represents an input image, and $y_i \in \{0,1\}^{C \times H \times W}$ is the corresponding ground-truth annotation with $C$ classes. 
The architecture of our KnowSAM framework is illustrated in \figref{Model}. \zh{Specifically, an input image is fed into two distinct subnets, $\mathcal{F}_A$ and $\mathcal{F}_B$. Following this, the segmentation outputs, $\hat{Y}_a$ and $\hat{Y}_b$, from the two subnets are subsequently input into a hybrid aggregation module, resulting in the composite map $\hat{Y}_f$. This map serves as the mask prompt for the SAM decoder.  Concurrently, the input image is passed through the SAM encoder to extract the image embedding, which is then processed by $\psi(\cdot)$ to derive the learnable feature prompt. Together with the mask prompt, these are provided to the SAM decoder to produce predictions. Subsequently, the outputs of multi-view co-training are refined by SAM predictions via a knowledge distillation process. Additionally, as highlighted by the yellow flow line in Fig.~\ref{Model}, we propose a UGDA strategy to strengthen the model's generalization capabilities. This strategy involves generating new hybrid images by substituting regions of low confidence, which are subsequently employed to retrain the model.}

\subsection{Multi-view Co-training}\label{sec:DCN}

The MC component comprises two distinct structural subnets with independent parameter updates, which are designed to correct each other through a simple aggregation strategy. In particular, the aggregation module focuses on addressing inherent uncertainty and areas of prediction inconsistency within the two sub-networks. To achieve this, we initially utilize the entropy map $\mathcal {H}$~\cite{wang2023s} as the uncertainty map, which can be defined as follows:
\begin{equation} 
\mathcal {H} = - \sum\nolimits_{c=0}^{C-1} \hat{Y_c}\times \log{\hat{Y_c}},
\end{equation}
where $\hat{Y_c}$ denotes the prediction for the $c$-th class of a sub-network. Hence, we can obtain the entropy maps $\mathcal {H}_a$ and $\mathcal {H}_b$ for the two sub-networks. Then, we utilize the XOR operation to determine areas of disagreement in predictions between the two sub-networks. The dissimilarity map, denoted as $\mathcal {M}$, is formalized by
\begin{equation} 
\mathcal {M} =  \mathcal {B}(\hat{Y_a}) \oplus \mathcal {B}(\hat{Y_b}),
\end{equation}
where $\hat{Y_a}$ and $\hat{Y_b}$ represent the predicted maps of the two sub-networks, respectively. $\mathcal {B}$ and $\oplus$ denote the binarization and XOR operation, respectively. To mitigate the impact of uncertainty and inconsistency predictions from the two subnets, we propose a Hybrid Aggregation Module (HAM) to fuse multi-view maps. These include the two prediction maps, entropy maps, and the dissimilarity map. Consequently, the aggregated prediction can be obtained by fusing these maps, which can be formulated by 
\begin{equation} 
\hat{Y}_f = \mathcal{F}_{\textup{CBL}}(\mathcal{F}_{\textup{CAT}}\{\mathcal{F}_{\textup{CBL}}(\mathcal {H}_a, \mathcal {H}_b), \mathcal{F}_{\textup{CBL}}(\mathcal {M}), \mathcal{F}_{\textup{CBL}}(\hat{Y}_a, \hat{Y}_b)\}),
\end{equation}
where $\mathcal{F}_{\textup{CBL}}$ and $\mathcal{F}_{\textup{CAT}}$ denote the Conv-BatchNorm-LeakyReLU and concatenation operations, respectively. Note that, the aggregated prediction maps are only supervised by the ground truth labels of labeled data. Therefore, the supervised loss can be formulated as follows:
\begin{equation} 
\mathcal {L}_{fuse} =  \mathcal {L}_{SEG}(\hat{Y}_f, Y^l),
\end{equation}
where $\mathcal {L}_{SEG}=\mathcal {L}_{DICE}+\mathcal {L}_{CE}$, with $\mathcal {L}_{DICE}$ and $\mathcal {L}_{CE}$ denoting the Dice loss and Cross-Entropy (CE) loss, respectively. ${Y^l}$ denotes ground truth labels of labeled data. 
Furthermore, we adopt the mutual supervision strategy as presented in~\cite{chen2021semi} to ensure the alignment between $\hat{Y}_a$ and $\hat{Y}_b$, generated by the two subnets. Specifically, $\hat{Y}_a$ acts as a pseudo-label for $\hat{Y}_b$, and vice versa, establishing mutual consistency between them. The mutual consistency loss is defined as follows:
\begin{equation} 
\mathcal {L}_{mut} =  \mathcal {L}_{CE}(\hat{Y}_a, \hat{Y}_b) + \mathcal {L}_{CE}(\hat{Y}_b, \hat{Y}_a).
\end{equation}

\subsection{Learnable Prompt for SAM}
\label{sec:Visual Prompting SAM}

We present a Learnable Prompt Strategy (LPS) to generate learnable feature prompt that are fed into the decoder of SAM. This effectively alleviates the significant performance degradation of SAM caused by inputting incorrect sparse information (\eg, containing only horizontal and vertical coordinate points). Specifically, as depicted in Fig.~\ref{Model}, we devise a lightweight decoder $\psi(\cdot)$ to generate dense prompts (embedding features), which can be represented as follows:
\begin{equation} 
\mathbf{P}_b = \psi(Z;{\Theta}_m),
\end{equation}
where $Z \in \mathbb{R}^{B \times D \times H \times W}$ denotes the feature map of input images extracted by SAM's image encoder, where $B, D, H$, and $W$  respectively represent the batch size, channel, width, and height. ${\Theta}_m$ represents the parameters of $\psi(\cdot)$. $\mathbf{P}_b\in\mathbb{R}^{B \times N_b\times L}$ denotes learnable feature prompt, with $N_b$ denoting the number of feature prompts. 
Moreover, the aggregated map from the two subnets is used to form the mask prompt. Therefore, the above process can be expressed as follows:
\begin{equation} 
\hat{Y}_s = \mathcal {F}_s(\mathbf{P}_b, \hat{Y}_f; {\Theta}_s),
\end{equation}
where $\mathcal {F}_s(\cdot)$ represents SAM's decoder, $\hat{Y}_s$ denotes the SAM's prediction. ${\Theta}_s$ represents the parameters of $\mathcal {F}_s(\cdot)$. Moreover, we introduce an adapter~\cite{wu2023medical} tuning strategy to the image encoder and decoder of the SAM model for fine-tuning, aligning the model with this new prompt approach. SAM is solely supervised by ground truth labels, thus the loss function is formulated by
\begin{equation} 
\mathcal {L}_{sam} =  \mathcal {L}_{SEG}(\hat{Y}_s, Y^l).
\end{equation}

\subsection{SAM-induced Knowledge Distillation}\label{sec:Knowledge Distillation}

We propose the SKD strategy to transfer useful knowledge from SAM to two subnets, thereby enhancing the intrinsic generalization capability of each subnet. More importantly, by introducing distillation learning, the overfitting of subnets to noise can be mitigated~\cite{zhong2023semi}. Following the approach of KD~\cite{hinton2015distilling}, we employ temperature calibrated softmax (T-S) to soften probability maps in the Kullback-Leibler divergence function, which can be expressed as:
\begin{equation} 
\hat{Y}^c_{T} = \frac{\exp(\hat{\mathbf{q}}_c / T)}{\sum_{c} \exp(\hat{\mathbf{q}}_c / T)},
\end{equation}
where $\hat{\mathbf{q}}_c$ is the logit prediction for class $c$, and $\hat{Y}^c_{T}$ is the soft probability value for class $c$. Besides, $T$ denotes a temperature parameter, and a high value can generate a softer probability distribution. After that, we utilize SAM as the teacher model and employ two sub-models as student models for knowledge distillation. 
Thus, the KD loss function can be expressed as follows:
\begin{equation} 
\mathcal {L}_{kd} = KL(\hat{Y}_{T}^{a}, \hat{Y}_{T}^{s}) + KL(\hat{Y}_{T}^{b}, \hat{Y}_{T}^{s}),
\end{equation}
where $\hat{Y}_T^a$, $\hat{Y}_T^b$ and $\hat{Y}_T^s$ denote the soft probability maps of two subnets and SAM, respectively. 
$KL(\cdot)$ is the Kullback-Leibler divergence function. 
Note that the gradient of $\mathcal {L}_{kd}$ is solely back-propagated to the two subnets, facilitating knowledge distillation from SAM to the subnets.

\begin{figure}[!t]
	\centering
	\begin{overpic}[width=0.5\textwidth]{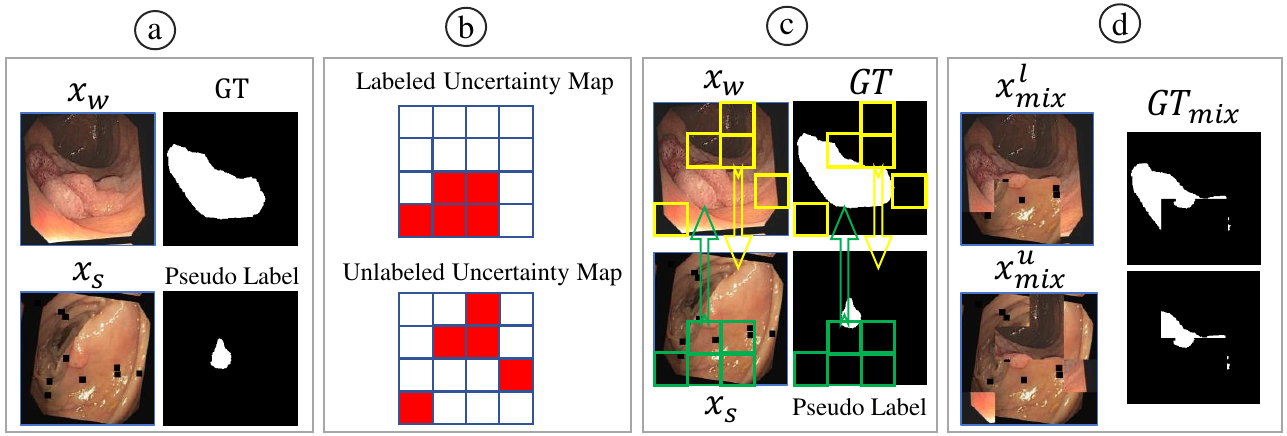}
    \end{overpic}\vspace{-0.15cm}
	\caption{{Pipeline of the proposed UGDA strategy. (a) Labeled data undergo weak augmentation with ground-truth labels, while unlabeled data receive strong augmentation with pseudo labels. (b) An uncertainty map is obtained from the aggregated prediction $\hat{Y}_f$, and the top five regions with the highest uncertainty are identified. (c) Interactive bidirectional copy-paste is employed based on the uncertainty map (a single region is depicted here for demonstration purposes). (d) Mixed images with labels are used as additional training samples.
 }}\vspace{-0.15cm}
    \label{UGDA}
\end{figure}

\subsection{Uncertainty-Guided Data Augmentation}

A data augmentation method, such as CutMix~\cite{yun2019cutmix}, is a straightforward and effective way to enhance model segmentation performance while bolstering model robustness. However, current copy-paste methods (\eg, BCP~\cite{bai2023bidirectional} and MiDSS~\cite{ma2024constructing}) are typically random, leading to instability. 
Therefore, we propose the UGDA strategy to address this issue. As shown in \figref{UGDA}, in the first step, we apply weak augmentation to labeled data and strong augmentation to unlabeled data, thus we can obtain as follows:
\begin{equation}
x_w = \mathcal {A}^W(x^l_i),~~x_s = \mathcal {A}^S(x^u_i),
\end{equation}
where $\mathcal {A}^W$ and  $\mathcal {A}^S$ represent weak and strong data augmentation, respectively. In the second step, the data are fed into the entire model to obtain the aggregated result $\hat{Y}_f$. The cutting positions are determined by calculating the entropy uncertainty~\cite{wang2023s} of $\hat{Y}_f$. Then, we divide the uncertainty map into $N$ patches, where $N$ is set to $16$. Moreover, we calculate the mean uncertainty value for each region using average pooling and select the top five patches with the highest uncertainty value. In the third step, \zh{we leverage the patch coordinates indicated by the uncertainty map to perform a cross copy-pasting operation between labeled and unlabeled data, thereby creating mixed images. 
Additionally, the outputs from SAM serve as the pseudo-labels for the unlabeled data. Consequently, this strategy not only enhances the diversity of the sample set but also mitigates the noise stemming from uncertain predictions}.


\subsection{Overall Loss Function}\label{sec:Overall Loss Function}

To effectively conduct distillation learning from SAM to our subnets, the training scheme comprises two components: the SAM loss and the subnets' loss. In this scenario, it is preferable to carry out inference using only the subnets. The details of the SAM loss can be found in Sec.~\ref{sec:Visual Prompting SAM}. Additionally, the loss of subnets includes supervised loss, KD loss, entropy loss, and mutual consistency loss. Initially, the supervised loss on labeled data can be formulated as follows:
\begin{equation} 
\mathcal {L}_{sup} =  \mathcal {L}_{SEG}(\hat{Y}_a, {Y^l}) + \mathcal {L}_{SEG}(\hat{Y}_b, {Y^l}) + \mathcal {L}_{fuse}.
\end{equation}
Then, minimizing entropy can enhance the confidence of model predictions. The entropy loss function is defined as follows:
\begin{equation} 
\begin{aligned}
& \mathcal {L}_{ent} =  -\frac{1}{H{\times}W}  \sum\nolimits_{k}^{H{\times}W} \sum\nolimits_{c=0}^{C} (\hat{Y}_a^c(k)\log(\hat{Y}_a^c(k)) \\ 
             & \qquad + \hat{Y}_b^c(k)\log(\hat{Y}_b^c(k))),
\end{aligned}
\end{equation}
where 
$\hat{Y}_a^c(k)$ and $\hat{Y}_b^c(k)$ are the predictions of two sub-networks for class $c$ at pixel $k$. 
Hence, the overall loss function for the subnets is as follows:
\begin{equation} 
\mathcal {L}_{total} =  \mathcal {L}_{sup} + \mathcal {L}_{kd} + \lambda_{e}\mathcal {L}_{ent} + \lambda_{m}\mathcal {L}_{mut},
\end{equation}
where $\lambda_{e}$ and $\lambda_{m}$ are coefficients for entropy loss and mutual consistency loss, respectively. It is noteworthy that there is no supervised loss for unlabeled data.



\begin{table*}[t!]
  \centering
  \scriptsize
  \small
  \renewcommand{\arraystretch}{1.0}
  \setlength\tabcolsep{3.8pt}
  \caption{Quantitative results on colonoscopy, ISIC-2018 and ultrasound thyroid nodule datatsets (``MC-Seg" denotes our multi-view co-training with UGDA).
  }\label{tab_tumor}\vspace{-0.15cm}
  \begin{tabular}{r|c|ccc|ccc|c|ccc|ccc}
  \hline
  \multicolumn{1}{c|}{\multirow{2}*{\textbf{Methods}}} &  & \multicolumn{3}{c|}{$10\%$ labeled} & \multicolumn{3}{c|}{$30\%$ labeled} &  & \multicolumn{3}{c|}{$10\%$ labeled} & \multicolumn{3}{c}{$30\%$ labeled}\\
  \cline{3-8}\cline{10-15}
   & & Dice(\%) & IoU(\%)  & 95HD & Dice(\%) & IoU(\%) & 95HD & & Dice(\%) & IoU(\%)  & 95HD & Dice(\%) & IoU(\%) & 95HD\\
    \hline
  MT~\cite{tarvainen2017mean} & \multirow{15}{*}{\begin{sideways}CVC-300\end{sideways}} & 56.64 & 48.90 & 4.69 & 78.27 & 68.69 & 3.86 & \multirow{15}{*}{\begin{sideways}CVC-ClinicDB\end{sideways}}  & 78.64 & 71.02 & 3.97 & 83.21 & 76.92 & \textcolor{blue}{3.50}\\
  UA-MT \cite{yu2019uncertainty} & & 40.15 & 32.46 & 4.97 & 80.95 & 70.77 & \textcolor{red}{3.19}  & & 74.51 & 67.14 & 4.05 & 78.89 & 72.61 & 3.68\\
  DTC~\cite{luo2021semi} & & 42.81 & 33.35 & 5.42 & 77.44 & 67.43 & 3.32  & & 67.82 & 57.96 & 4.61 & 74.28 & 66.21 & 3.97\\
  URPC~\cite{luo2022semi} & & 58.83 & 50.81 & 4.30 & 77.05 & 66.75 & 3.65  & & 76.42 & 69.33 & 3.78 & 80.70 & 74.58 & 3.64\\
  MC-Net~\cite{wu2021semi} & & 69.89 & 61.58 & 3.93 & 81.36 & 72.11 & 3.33  & & 75.21 & 67.87 & 4.10 & 83.53 & 76.78 & 3.56\\
  MC-Net+~\cite{wu2022mutual} & & 70.08 & 62.43 & 4.09 & 81.01 & 70.93 & 3.41  & & 78.23 & 70.83 & 3.94 & 83.77 & {77.43} & 3.58\\
  MCF~\cite{wang2023mcf} & & 67.58 & 57.85 & 3.91 & 79.51 & 69.03 & 3.27  & & 70.83 & 62.41 & 4.06 & 80.41 & 74.50 & 3.63\\
  CDMA~\cite{zhong2023semi} & & 59.44 & 48.93 & 4.34 & 66.38 & 52.84 & 4.77  & & 74.18 & 65.31 & 4.09 & 80.81 & 72.79 & 3.81\\
  CauSSL~\cite{miao2023caussl} & & 64.93 & 55.06 & 4.33 & 71.92 & 62.02 & 3.71  & & 76.23 & 68.44 & 3.99 & 75.18 & 68.08 & 3.89\\
  BS-Net~\cite{he2023bilateral} & & 65.46 & 56.31 & 4.09 & 80.53 & 70.99 & 3.33  & & 74.83 & 67.44 & 4.03 & 80.32 & 74.15 & 3.62\\
  BCP~\cite{bai2023bidirectional} & & {78.33} & {68.94} & {3.89} & {82.51} & 73.07 & 3.38  & & \textcolor{blue}{82.84} & {75.76} & \textcolor{blue}{3.64} & \textcolor{blue}{84.21} & 76.64 & 3.67\\
  CAML~\cite{gao2023correlation} & & 70.66 & 63.40 & 3.99 & 82.25 & {74.14} & {3.27}  & & 77.84 & 71.71 & 3.71 & 80.91 & 74.93 & 3.67\\
  \hline
  \textbf{MC-seg} (Ours) & & \textcolor{blue}{82.81} & \textcolor{blue}{74.25} & \textcolor{blue}{3.49} & \textcolor{blue}{84.09} & \textcolor{blue}{76.17} & 3.35 & & {82.27} & \textcolor{blue}{75.82} & {3.78} & {83.95} & \textcolor{blue}{78.15} & {3.66}\\ 
\textbf{KnowSAM} (Ours) & & \textcolor{red}{82.86} & \textcolor{red}{75.19} & \textcolor{red}{3.46} & \textcolor{red}{84.28} & \textcolor{red}{76.98} & \textcolor{blue}{3.20} & & \textcolor{red}{83.36} & \textcolor{red}{77.39} & \textcolor{red}{3.48} & \textcolor{red}{85.09} & \textcolor{red}{79.82} & \textcolor{red}{3.44}\\ 
  \hline\hline

  MT~\cite{tarvainen2017mean} & \multirow{15}{*}{\begin{sideways}CVC-ColonDB\end{sideways}} & 47.67 & 39.30 & 5.71 & 63.03 & 54.63 & 4.80 & \multirow{15}{*}{\begin{sideways}ETIS\end{sideways}}  & 33.64 & 26.78 & 5.41 & 51.06 & 42.86 & 4.63\\
  UA-MT \cite{yu2019uncertainty} & & 41.34 & 33.49 & 5.72 & 59.43 & 51.30 & 4.74  & & 31.84 & 25.59 & 4.96 & 41.18 & 34.80 & 4.64\\
  DTC~\cite{luo2021semi} & & 35.35 & 27.17 & 6.10 & 54.75 & 46.18 & 5.05 & & 26.74 & 20.48 & 6.05 & 44.83 & 37.72 & 4.40\\
  URPC~\cite{luo2022semi} & & 46.84 & 38.98 & 5.55 & 60.00 & 51.89 & 4.78  & & 38.98 & {32.92} & {4.78} & 45.00 & 36.75 & 4.89\\
  MC-Net~\cite{wu2021semi} & & 55.69 & 46.53 & 5.24 & 63.65 & 55.38 & 4.71  & & 39.88 & 33.15 & 4.81 & 49.51 & 41.69 & 4.74\\
  MC-Net+~\cite{wu2022mutual} & & 52.36 & 44.52 & 5.47 & 65.94 & 57.33 & 4.62  & & 41.71 & 33.92 & 5.20 & 45.07 & 38.67 & 4.69\\
  MCF~\cite{wang2023mcf} & & 48.72 & 40.60 & 5.25 & 63.24 & 55.05 & 4.64  & & 31.13 & 25.81 & 4.84 & 48.46 & 41.28& \textcolor{blue}{4.33}\\
    CDMA~\cite{zhong2023semi} & & 44.88 & 35.72 & 5.77 & 59.39 & 49.10 & 5.20  & & 37.23 & 28.85 & 4.87 & 38.29 & 29.37 & 6.07\\
  CauSSL~\cite{miao2023caussl} & & 48.51 & 40.89 & 5.24 & 57.84 & 49.36 & 4.95  & & 30.93 & 24.96 & 5.18 & 32.97 & 27.41 & 5.42\\
  BS-Net~\cite{he2023bilateral} & & 43.88 & 36.57 & 5.43 & 55.89 & 48.15 & 4.87  & & 31.77 & 26.26 & 5.28 & 38.08 & 31.65 & 4.91\\
  BCP~\cite{bai2023bidirectional} & & {59.91} & 51.31 & 4.93 & {68.11} & {58.91} & \textcolor{blue}{4.61}  & & {46.10} & {37.52} & 4.95 & {55.28} & 46.09 & 4.52\\
  CAML~\cite{gao2023correlation} & & {59.86} & {52.96} & \textcolor{blue}{4.90} & 64.23 & 56.69 & 4.64  & & 33.27 & 28.94 & {4.63} & 54.80 & {47.03} &  {4.36}\\
  \hline
  \textbf{MC-seg} (Ours) & & \textcolor{blue}{64.15} & \textcolor{blue}{55.92} & {5.02} & \textcolor{blue}{68.71} & \textcolor{blue}{60.30} & {4.76} & & \textcolor{blue}{47.55} & \textcolor{blue}{41.40} & \textcolor{blue}{4.63} & \textcolor{blue}{56.04} & \textcolor{blue}{48.89} & {4.43}\\ 
\textbf{KnowSAM} (Ours) & & \textcolor{red}{69.76} & \textcolor{red}{62.14} & \textcolor{red}{4.51} & \textcolor{red}{70.37} & \textcolor{red}{62.53} & \textcolor{red}{4.57} & & \textcolor{red}{52.62} & \textcolor{red}{45.89} & \textcolor{red}{4.40} & \textcolor{red}{58.69} & \textcolor{red}{52.40} & \textcolor{red}{4.07}\\ 
  \hline\hline
  
  MT~\cite{tarvainen2017mean} & \multirow{15}{*}{\begin{sideways}Kvasir\end{sideways}} & 73.99 & 64.19 & 5.37 & 83.74 & 75.89 & 4.53 & \multirow{15}{*}{\begin{sideways}ISIC-2018\end{sideways}}  & 80.86 & 72.04 & 5.32 & 83.70 & 75.07 & 4.96\\
  UA-MT \cite{yu2019uncertainty} & & 73.15 & 64.12 & 5.06 & 83.71 & 75.95 & 4.51  & & 80.15 & 71.48 & 5.37 & 85.08 & 77.19 & 4.90\\
  DTC~\cite{luo2021semi} & & 66.79 & 56.00 & 5.63 & 77.73 & 68.78 & 4.73  & & 80.02 & 71.74 & 5.19 & 83.53 & 75.21 & 4.89\\
  URPC~\cite{luo2022semi} & & 79.14 & 70.53 & 4.65 & 84.47 & 76.63 & 4.34  & & 79.38 & 70.36 & 5.37 & 83.18 & 74.85 & 4.91\\
  MC-Net~\cite{wu2021semi} & & 78.03 & 69.21 & 4.89 & 85.51 & 78.07 & 4.46  & & 80.98 & 72.03 & 5.31 & 83.21 & 74.42 & 4.98\\
  MC-Net+~\cite{wu2022mutual} & & 78.57 & 69.59 & 4.84 & 86.18 & 79.57 & \textcolor{red}{4.21}  & & 80.72 & 71.75 & 5.37 & 84.37 & 76.50 & 4.85\\
  MCF~\cite{wang2023mcf} & & 74.01 & 64.65 & 4.99 & 83.25 & 75.42 & 4.46  & & 80.76 & 71.88 & 5.09 & 83.50 & 75.13 & 4.86\\
  CDMA~\cite{zhong2023semi} & & 74.98 & 65.06 & 5.07 & 82.43 & 73.47 & 4.63  & & 76.70 & 66.58 & 5.84 & 83.27 & 74.88 & \textcolor{red}{4.78}\\
  CauSSL~\cite{miao2023caussl} & & 76.48 & 67.61 & 4.93 & 81.03 & 73.20 & 4.67  & & 79.73 & 71.19 & 5.34 & 84.53 & 75.88 & 4.93\\
  BS-Net~\cite{he2023bilateral} & & 77.18 & 68.22 & 4.99 & 82.50 & 74.98 & 4.45  & & 81.70 & 73.68 & 5.05 & 84.22 & 75.83 & 4.85\\
  BCP~\cite{bai2023bidirectional} & & 79.64 & 70.44 & 4.85 & 85.99 & 78.52 & 4.56  & & {83.98} & {75.15} & 5.00 & {85.35} & {77.56} & {4.84}\\
  CAML~\cite{gao2023correlation} & & {80.66} & {73.27} & \textcolor{blue}{4.58} & \textcolor{blue}{87.25} & \textcolor{blue}{80.58} & {4.22}  & & 81.89 & 73.66 & \textcolor{blue}{4.96} & 83.65 & 75.57 & 4.86\\
  \hline
  \textbf{MC-seg} (Ours) & & \textcolor{blue}{84.47} & \textcolor{blue}{77.00} & {4.69} & {87.20} & {80.53} & {4.47} & & \textcolor{blue}{84.44} & \textcolor{blue}{76.36} & {5.09} & \textcolor{blue}{86.06} & \textcolor{blue}{77.84} & {5.04}\\ 
\textbf{KnowSAM} (Ours) & & \textcolor{red}{85.98} & \textcolor{red}{79.25} & \textcolor{red}{4.41} & \textcolor{red}{88.74} & \textcolor{red}{82.93} & \textcolor{blue}{4.22} & & \textcolor{red}{86.51} & \textcolor{red}{78.49} & \textcolor{red}{4.86} & \textcolor{red}{87.22} & \textcolor{red}{79.72} & \textcolor{blue}{4.81}\\ 
  \hline

  MT~\cite{tarvainen2017mean} & \multirow{15}{*}{\begin{sideways}DDTI\end{sideways}} & 31.59 & 24.32 & 6.79 & 47.62 & 38.17 & 6.24 & \multirow{15}{*}{\begin{sideways}TN3K\end{sideways}}  & 73.09 & 62.95 & 4.96 & 78.88 & 69.71 & 4.53 \\
  UA-MT \cite{yu2019uncertainty} & & 34.96 & 27.31 & 6.73 & 49.76 & 39.85 & 6.07  & & 74.73 & 64.23 & 4.83 & 78.42 & 69.16 & 4.53 \\
  DTC~\cite{luo2021semi} & & 27.13 & 19.67 & 7.35 & 38.53 & 28.05 & 6.87  & & 62.67 & 50.50 & 5.61 & 64.48 & 51.66 & 5.74 \\
  URPC~\cite{luo2022semi} & & 31.37 & 24.23 & 6.88 & 56.01 & 45.26 & 5.92  & & 74.42 & 64.17 & 4.88 & 79.19 & 70.00 & 4.56 \\
  MC-Net~\cite{wu2021semi} & & 29.42 & 23.12 & 7.15 & 45.36 & 36.60 & 6.20  & & 71.45 & 60.72 & 5.28 & 78.04 & 68.72 & 4.65 \\
  MC-Net+~\cite{wu2022mutual} & & 41.14 & 32.34 & 6.52 & 47.26 & 38.61 & 6.09  & & 75.05 & 65.13 & 5.10 & 79.70 & 70.26 & 4.50 \\
  MCF~\cite{wang2023mcf} & & {57.69} & {46.84} & \textcolor{blue}{5.71} & 55.09 & 44.08 & 5.97  & & 74.50 & 64.46 & 4.79 & 78.04 & 68.79 & 4.43 \\
    CDMA~\cite{zhong2023semi} & & 46.42 & 36.17 & 6.28 & 54.68 & 42.48 & 6.34  & & 72.97 & 61.54 & 5.12 & 76.24 & 65.60 & 4.81\\
  CauSSL~\cite{miao2023caussl} & & 41.88 & 32.76 & 6.52 & 46.12 & 36.51 & 6.22  & & 70.89 & 60.22 & 5.17 & 73.93 & 63.63 & 4.97\\
  BS-Net~\cite{he2023bilateral} & & 31.41 & 24.82 & 6.74 & 40.55 & 32.42 & 6.33  & & 73.70 & 64.22 & 4.77 & 79.25 & 70.60 & \textcolor{red}{4.39}\\
  BCP~\cite{bai2023bidirectional} & & 57.45 & 44.78 & 5.97 & 52.30 & 41.00 & 6.10  & & 75.36 & 64.65 & 5.04  & 76.70 & 66.59 & 4.82 \\
  CAML~\cite{gao2023correlation} & & 42.45 & 34.50 & 6.33 & {55.94} & {46.28} & \textcolor{blue}{5.73}  & & {77.78} & {68.80} & \textcolor{red}{4.47} & {80.15} & {71.30} & \textcolor{blue}{4.40} \\
  \hline
  \textbf{MC-seg} (Ours) & & \textcolor{blue}{58.20} & \textcolor{blue}{47.80} & {6.13} & \textcolor{blue}{58.16} & \textcolor{blue}{47.63} & {6.18} & & \textcolor{blue}{78.57} & \textcolor{blue}{69.42} & {4.88} & \textcolor{blue}{80.18} & \textcolor{blue}{71.50} & {4.65}\\ 
\textbf{KnowSAM} (Ours) & & \textcolor{red}{63.57} & \textcolor{red}{52.98} & \textcolor{red}{5.68} & \textcolor{red}{64.89} & \textcolor{red}{53.92} & \textcolor{red}{5.67} & & \textcolor{red}{81.19} & \textcolor{red}{72.27} & \textcolor{blue}{4.52}  & \textcolor{red}{81.21} & \textcolor{red}{72.18} & {4.67}\\ 
  \hline
  \end{tabular}
\end{table*}

\section{EXPERIMENTS AND RESULTS}\label{sec:Experiments}

\subsection{Datasets}

\textbf{Colonoscopy Datasets}: Five colonoscopy datasets are used for polyp segmentation, including ETIS~\cite{silva2014toward}, CVC-ClinicDB~\cite{bernal2015wm}, CVC-ColonDB~\cite{tajbakhsh2015automated}, CVC-300~\cite{vazquez2017benchmark}, and Kvasir~\cite{jha2020kvasir}. Following the previous setting~\cite{fan2020pranet}, a total of $1,450$ images are selected from the CVC-ClinicDB and Kvasir datasets to form the training set, while all remaining $798$ images are used for testing. 

\textbf{Ultrasound Thyroid Nodule Datasets}: Three ultrasound datasets are used, namely TN3K \cite{gong2023thyroid}, TG3K \cite{wunderling2017comparison}, and DDTI \cite{pedraza2015open}. The TN3K dataset consists of $3,493$ ultrasound images, with $2,879$ images for training and $614$ ones for testing. The DDTI dataset contains $637$ ultrasound images captured using the same device. The TG3K dataset includes $3,585$ ultrasound images collected from $16$ ultrasound videos. Following the experimental setup described in \cite{gong2023thyroid}, we combine $2,879$ images from TN3K and $3,585$ images from TG3K to form the training and validation datasets, while $614$ images from TN3K and $637$ images from DDTI are formed the test set.

\textbf{ISIC Dataset}: The ISIC-2018 \cite{codella2019skin} dataset,  provided by the International Skin Imaging Collaboration (ISIC), is derived from a substantial collection of dermoscopy images. It comprises $2,594$ images for training and $1,000$ images for testing. 

\textbf{ACDC Dataset}: The ACDC dataset~\cite{bernard2018deep}, introduced during the MICCAI 2017 challenge, includes four classes, (\ie, background, right ventricle, left ventricle, and myocardium). It comprises imaging scans from $100$ patients. Following the data partitioning strategy outlined in~\cite{bai2023bidirectional}, we split the dataset into $70$ scans for training, $10$ for validation, and $20$ for testing.

\zh{\textbf{BCSS Dataset}: The Breast Cancer Semantic Segmentation (BCSS) dataset~\cite{amgad2019structured} contains $151$ pathological images of varying sizes. In this study, we utilize an overlap-based cropping method to produce a dataset of $3,888$ images, each with a resolution of $1024 \times 1024$ pixels. This choice of dimension was strategic, aimed at ensuring that each image captures a comprehensive amount of tumor-related details.  Our focus is exclusively on the tumor category, with all other regions considered as background. These collected images are divided into training, validation, and test sets in the proportions of $70\%$, $10\%$, and $20\%$, respectively.}

\begin{figure*}
	\centering
	\begin{overpic}[width=0.96\textwidth]{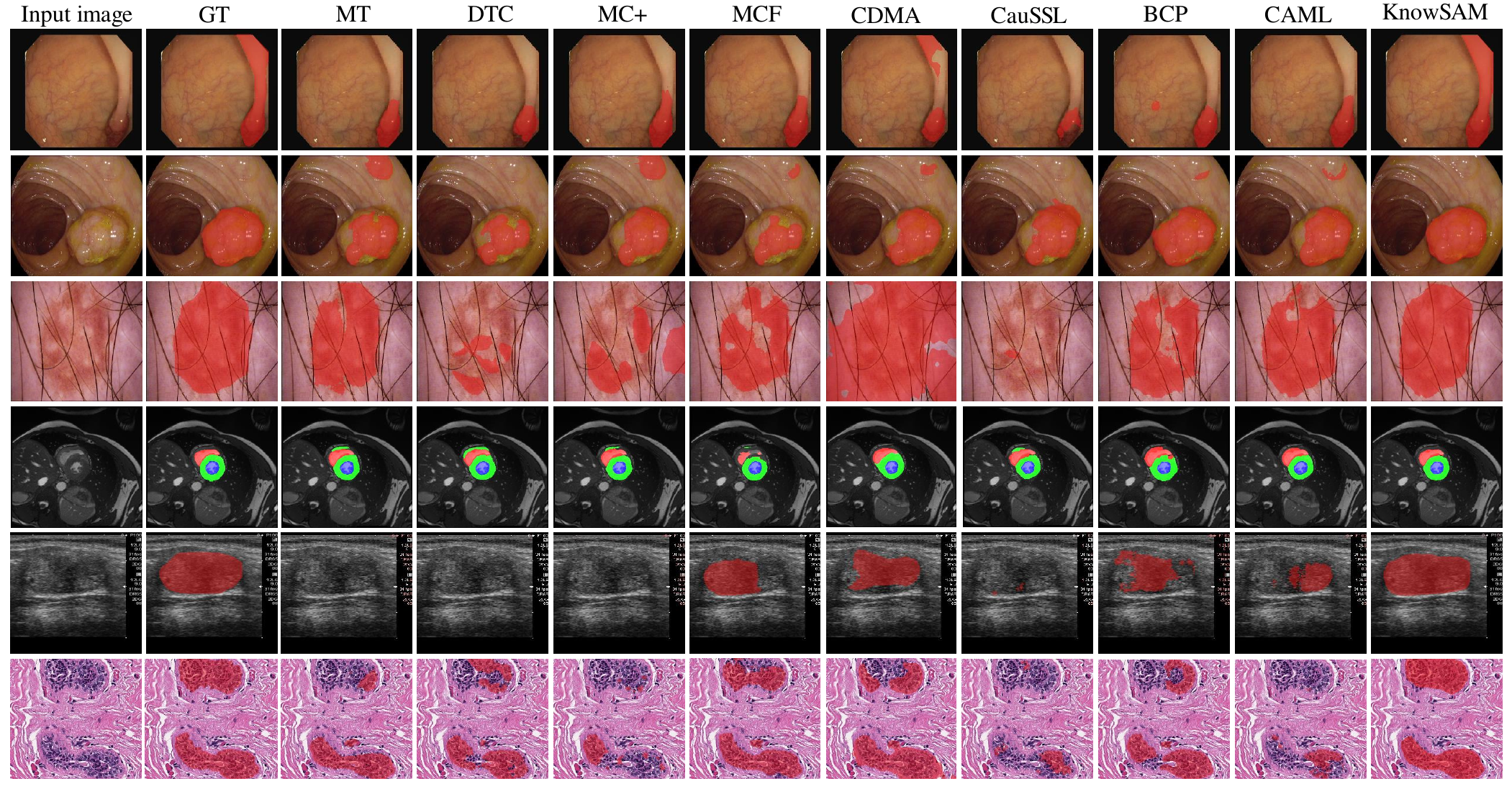}
    \end{overpic}\vspace{-0.25cm}
	\caption{{\zh{Visual results of different methods on five segmentation tasks. The comparison methods include MT and BCP, which utilize teacher-student architectures, while DTC, MC+, MCF, CDMA, CauSSL, and our method are based on consistency learning architectures. The first two rows correspond to endoscopic images, while the remaining rows represent results on the ISIC-2018, ACDC, ultrasound, and pathological image datasets, respectively.}
 } 
	}
    \label{results_1}\vspace{-0.35cm}
\end{figure*}

\subsection{Experimental Setup}

\subsubsection{Implementation Details} 
All experiments are conducted in a PyTorch 1.8.1 environment with CUDA 11.2 and an NVIDIA 3090 GPU. The two subnets in the study are U-Net~\cite{ronneberger2015u} and V-Net~\cite{milletari2016v}, and the SAM uses the base version. For SAM, the Adam optimizer is employed with an initial learning rate set to $0.0001$. Other subnets use the SGD optimizer with a momentum of $0.9$, a weight decay of $0.0001$, and an initial learning rate of $0.01$. Batch size and the number of iterations are set to $24$ and 50k. The input size of all images is fixed at $256\times 256$.
Moreover, we use a Gaussian warm-up function $\lambda_{m}(t) = \beta * e^{-5(1-t/t_{max})^2}$ in consistency loss, where $\beta$ is set to $1$. $\lambda_{e}$ is empirically set to $0.9$. 
Note that in the training without SAM, we utilize the fused predictions $\hat{Y}_f$ as pseudo-labels for unlabeled data. During inference, the $\hat{Y_f}$ serves as the final segmentation results. For all tasks, $10\%$ or $30\%$ are randomly selected from the training set as labeled data, while the remaining ones are adopted as unlabeled data.

\subsubsection{Evaluation metrics} 
For 2D datasets, we employ three commonly used evaluation metrics, namely the Dice coefficient (Dice), Intersection over Union (IoU), and 95\% Hausdorff Distance (95HD). For 3D data, we introduce an additional metric, \ie, Average Surface Distance (ASD).

\subsection{Comparison with State-of-the-art Methods}

We compare the proposed model with 12 SOTA semi-supervised methods, namely MT~\cite{tarvainen2017mean}, UA-MT~\cite{yu2019uncertainty}, DTC~\cite{luo2021semi}, MC-Net~\cite{wu2021semi}, MC-Net+~\cite{wu2022mutual}, URPC~\cite{luo2022semi}, MCF~\cite{wang2023mcf}, BCP~\cite{bai2023bidirectional}, and CAML~\cite{gao2023correlation}. \zh{For
a fair comparison, consistent data preprocessing is applied to all the models, and the open-source code and hyperparameters are used for each method}.


\subsubsection{Results on Colonoscopy Datasets} 

Table \ref{tab_tumor} shows the quantitative results of various semi-supervised methods on five colonoscopy and the ISIC datasets. In the polyp segmentation task, we utilize two ratios of labeled data: $10\%$ ($145$ labeled images) and $30\%$ ($435$ labeled images) in the comparison experiments. 
Without using SAM distillation and utilizing only $10\%$ labeled data for training, our semi-supervised method achieves satisfactory results. Moreover, our method with SAM distillation obtains significant improvements in the dice coefficient on unseen datasets such as CVC-ColonDB and ETIS.

\begin{table}[t!]
  \centering
  \scriptsize
  \renewcommand{\arraystretch}{1.0}
  \setlength\tabcolsep{4.5pt}
  \caption{Quantitative results on the ACDC dataset.
  }\label{tab_ACDC}\vspace{-0.15cm}
\begin{tabular}{r|c | c|c c c c}
\hline
{\multirow{2}{*}{Methods}} & \multicolumn{2}{c|}{Scans used}                         & \multicolumn{4}{c}{Metrics} \\
\cline{2-7}
& Labeled                   & Unlabeled                  & Dice (\%)  & IoU (\%)  & HD95  & ASD  \\
\hline
MT~\cite{tarvainen2017mean} & \multirow{14}{*}{7(10\%)} & \multirow{14}{*}{63(90\%)} & 81.89  & 71.10  &  12.48 & 3.77 \\
UA-MT~\cite{yu2019uncertainty} &  &  & 81.65 & 70.64 &  6.88   & 2.03  \\
DTC~\cite{luo2021semi}     &  &  & 84.29  & 73.92 &  12.81 & 4.01  \\
URPC~\cite{luo2022semi}  & &   &   83.10    &  72.41    &   4.84    &  1.53    \\
MC-Net~\cite{wu2021semi}  &    &   &  86.44     &  77.04    &   5.50    &   1.85   \\
MC-Net+~\cite{wu2022mutual}    &   &  &  85.30     &    75.76  &   12.11    &    3.27  \\
MCF~\cite{wang2023mcf} &  &  &  83.23   &  72.93   &   12.79    &  3.40   \\
CDMA~\cite{zhong2023semi} & & & 83.77 & 73.00 & 4.94  & 1.44 \\
CauSSL~\cite{miao2023caussl} & & & 86.80 & 77.48 & 5.73 &1.83 \\
BS-Net~\cite{he2023bilateral} & & & 77.51 & 65.38 & 17.49 & 4.84 \\
BCP~\cite{bai2023bidirectional} &   &  &   88.84    &   80.62   &   3.98    &  1.17    \\
CAML~\cite{gao2023correlation}   &    &  &  81.90     &  70.60    &   6.46    &  1.52    \\
\hline
\textbf{MC-seg} (Ours)  &   &   &     \textcolor{blue}{89.08}  &  \textcolor{blue}{81.00}    &  \textcolor{blue}{1.35}    &   \textcolor{blue}{0.37}   \\
\textbf{KnowSAM} (Ours)   &   &   &     \textcolor{red}{89.56}  &  \textcolor{red}{81.66}    &  \textcolor{red}{1.28}    &   \textcolor{red}{0.36}   \\
\hline\hline
MT~\cite{tarvainen2017mean} & \multirow{14}{*}{21(30\%)} & \multirow{14}{*}{49(70\%)} &   87.06  &  77.94 &  11.17   & 2.95\\
UA-MT~\cite{yu2019uncertainty} &   &  &  87.75  &  78.98  &  4.78  &   1.63   \\
DTC~\cite{luo2021semi}     & &  &  85.75     &  75.99    &   5.08    &    1.67  \\
URPC~\cite{luo2022semi}  &   &   &   88.36    &   79.96   &   3.04    &  0.77    \\
MC-Net~\cite{wu2021semi}  & &  &    88.04   &    81.01  &    3.54   &  1.82    \\
MC-Net+~\cite{wu2022mutual}    &  &  &  88.39 &   80.08  &  6.38    &  1.60      \\
MCF~\cite{wang2023mcf} &  &   &  86.22     &  76.73    &   7.42    &  1.99    \\
CDMA~\cite{zhong2023semi} & & & 84.68 & 74.46 & 6.27 & 1.93 \\
CauSSL~\cite{miao2023caussl} & & & 85.11 & 75.00 & 9.80 & 2.99 \\
BS-Net~\cite{he2023bilateral} & & &  {89.31} & {81.24} & {2.10} & 0.68 \\
BCP~\cite{bai2023bidirectional} &  &   &  88.61     & 80.12     &  4.71     & 1.25     \\
CAML~\cite{gao2023correlation}   & &   &   86.85    &   77.63   &   5.08    &  1.55    \\
\hline
\textbf{MC-seg} (Ours)    &   &   &   \textcolor{blue}{90.22}    &    \textcolor{blue}{82.76}  &   \textcolor{red}{1.21}    & \textcolor{blue}{0.30}     \\
\textbf{KnowSAM} (Ours)   &   &   &     \textcolor{red}{91.13}  &  \textcolor{red}{84.11}    &   \textcolor{blue}{1.29}    &   \textcolor{red}{0.30}   \\
  \hline
  \end{tabular}\vspace{-0.15cm}
\end{table}

\subsubsection{Results on Ultrasound Datasets}


Table \ref{tab_tumor} shows the quantitative results across three ultrasound datasets. Due to the complex background in ultrasound images, incomplete segmentation is a common challenge. With only $10\%$ labeled data, our method achieves a $0.51\%$ dice coefficient higher than the second-best method MCF on the unseen DDTI dataset. On the TN3K dataset, the dice coefficient increases from $77.78\%$ to $78.57\%$ when compared to CAML with 10\% training data. More importantly, introducing SAM distillation further enhances the performance comprehensively, indicating KnowSAM's robust generalization across different scenarios.

\subsubsection{Results on ISIC-2018} As shown in Table \ref{tab_tumor}, our method improves the dice coefficient from $83.98\%$ to $84.44\%$ compared to BCP on the ISIC-2018 dataset with $10\%$ training data. By incorporating SAM distillation into our model, the Dice coefficient further increases from $84.44\%$ to $86.51\%$. Moreover, as shown in Fig. \ref{results_1}, our model completely segmentation tumor regions than other comparison methods.

\subsubsection{Results on ACDC} We further validate the proposed model on the multi-class task using the ACDC dataset. As shown in Table \ref{tab_ACDC} and Fig. \ref{results_1}, KnowSAM consistently outperforms other semi-supervised methods. With only $30\%$ labeled data (21 labeled samples), the Dice coefficient increases from $88.61\%$ to $90.22\%$ compared to the second-best method. Moreover, our method incorporating SAM distillation further improves the Dice coefficient to $91.13\%$, underscoring the effectiveness of our semi-supervised model and the robustness of the proposed KnowSAM framework.

\subsubsection{Results on BCSS} \zh{Moreover, we conduct comparison experiments on the pathology image segmentation task. As presented in Table \ref{tab_BCSS}, with only $10\%$ labeled data, our MC-seg achieves a Dice coefficient of $68.11\%$, which outperforms the two state-of-the-art methods, CDMA ($65.58\%$) and BCP ($66.00\%$). More notably, by integrating the SAM-induced knowledge distillation strategy, our KnowSAM model records a substantial improvement, achieving a Dice coefficient of $72.69\%$ and outperforming all other methods. This trend continues even with $30\%$ labeled data. This outcome underscores the KnowSAM framework's capacity to preserve model generalization and precision in identifying foreground target boundaries, even in challenging tasks such as pathological segmentation.}


\subsection{Ablation Study}

\subsubsection{Effectiveness of Different Views in MC}

\begin{figure}[!h]
	\centering
	\begin{overpic}[width=0.46\textwidth]{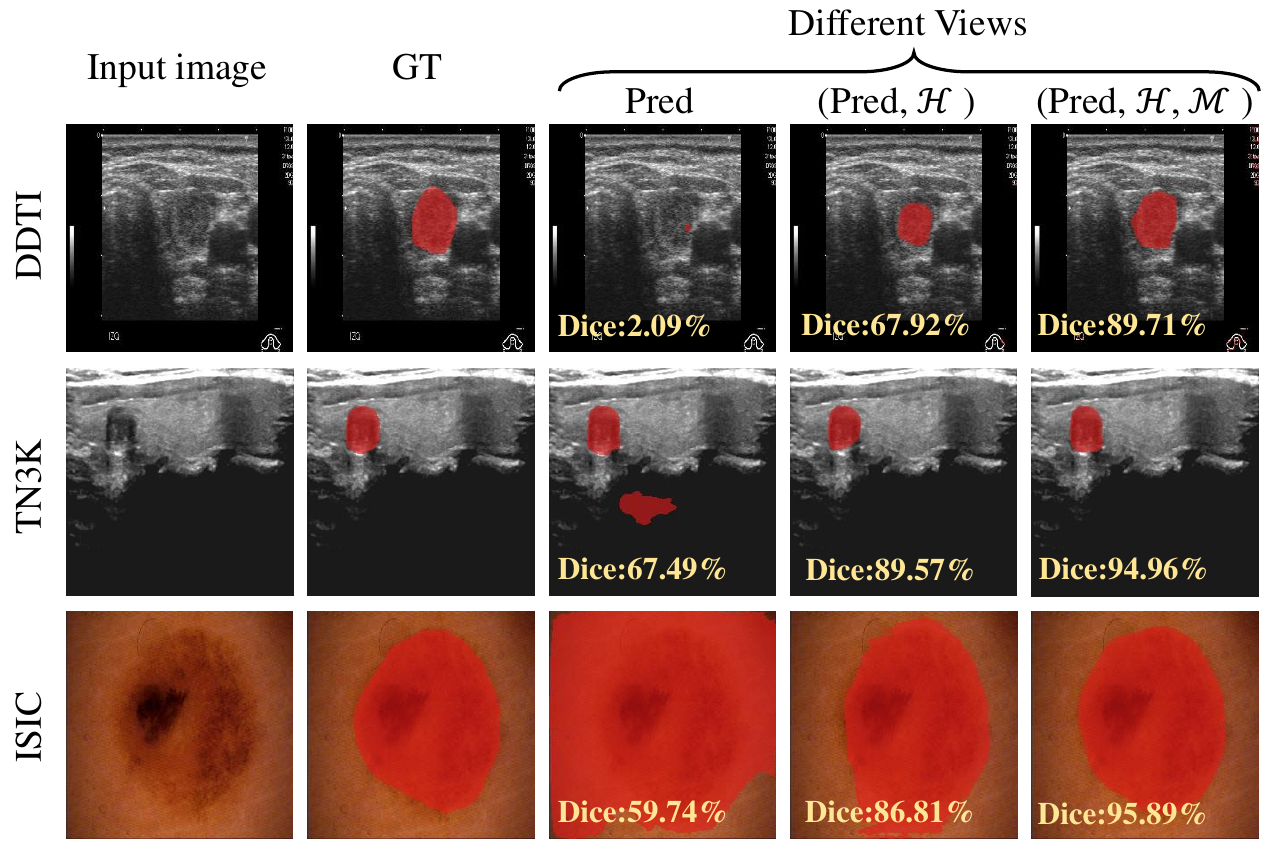}
    \end{overpic}\vspace{-0.25cm}
	\caption{{Visualization results of progressively adding different view information without SAM distillation using $30\%$ labeled data.} 
	}
    \label{Abi_views}\vspace{-0.25cm}
\end{figure}

In the MC component, we utilized the proposed HAM to fuse different maps, including the predictions of two sub-networks, entropy uncertainty maps ($\mathcal {H}$), and dissimilarity maps ($\mathcal {M}$), to produce reliable segmentation results. 
\zh{Notably, the uncertainty perspective effectively addresses the network's inherent cognitive limitations, enabling the model to identify and accommodate its uncertainties throughout the segmentation process. In addition, the dissimilarity perspective plays a key role in mitigating cognitive biases that may exist between the two subnets.} 
Table \ref{tab_abi_views} demonstrates the effectiveness of different views within the HAM strategy. \figref{Abi_views} presents the visualization results without SAM distillation.
Among these views, entropy uncertainty reflects the model's internal understanding of the data, while the inconsistent region identifies controversial areas requiring additional supervision. 
\zh{When training with $30\%$ labeled samples on the ACDC dataset, incorporating $\mathcal {H}$ view information increases the Dice coefficient from $88.63\%$ to $89.18\%$, with HD95 decreasing from $5.14$ to $4.06$. Furthermore, integrating $\mathcal {M}$ view information raises the Dice coefficient to $90.22\%$, and HD95 decreases to $1.21$. This highlights that entropy uncertainty and inconsistent regions can supplement the model's cognitive capabilities effectively, enhancing the semi-supervised segmentation performance.}

\begin{table}[t!]
  \centering
  \scriptsize
  \renewcommand{\arraystretch}{1.0}
  \setlength\tabcolsep{4.8pt}
  \caption{\zh{Quantitative results on the BCSS dataset}.
  }\label{tab_BCSS}\vspace{-0.15cm}
\begin{tabular}{r|c | c|c c c }
\hline
{\multirow{2}{*}{Methods}} & \multicolumn{2}{c|}{Scans used}                         & \multicolumn{3}{c}{Metrics} \\
\cline{2-6}
& Labeled                   & Unlabeled                  & Dice (\%)  & IoU (\%)  & HD95    \\
\hline
MT~\cite{tarvainen2017mean} & \multirow{14}{*}{7(10\%)} & \multirow{14}{*}{63(90\%)} & 63.93  & 50.75  &  9.04  \\
UA-MT~\cite{yu2019uncertainty} &  &  & 63.05  & 49.65  &  9.28   \\
DTC~\cite{luo2021semi}     &  & & 63.60  & 50.19  &  9.36  \\
URPC~\cite{luo2022semi}  & &   & 63.93  & 50.15  &  9.23    \\
MC-Net~\cite{wu2021semi}  &    &   & 63.67  & 50.12  &  9.21    \\
MC-Net+~\cite{wu2022mutual}  &  &   & 62.46  & 49.01  &  9.15  \\
MCF~\cite{wang2023mcf} &  &  & 65.52  & 52.14  &  8.98    \\
CDMA~\cite{zhong2023semi} & & & 65.58  & 52.25  &  8.92 \\
CauSSL~\cite{miao2023caussl} & & & 62.88  & 49.35  &  9.45 \\
BS-Net~\cite{he2023bilateral} & & & 64.47  & 52.37  &  \textcolor{blue}{8.28}  \\
BCP~\cite{bai2023bidirectional} &   &  & 66.00  & 53.11  &  8.67    \\
CAML~\cite{gao2023correlation}   &    &  & 65.56  & 52.64  &  8.93   \\
\hline
\textbf{MC-seg} (Ours)  &   &   & \textcolor{blue}{68.11}  & \textcolor{blue}{55.97}  &  8.32 \\
\textbf{KnowSAM} (Ours)   &   &   & \textcolor{red}{72.69}  &  \textcolor{red}{61.09}  &  \textcolor{red}{7.77} 
\\
\hline\hline
MT~\cite{tarvainen2017mean} & \multirow{14}{*}{21(30\%)} & \multirow{14}{*}{49(70\%)} & 68.87  & 56.15  &  8.53 \\
UA-MT~\cite{yu2019uncertainty} &   &  & 67.00  & 54.28  &  8.81 \\
DTC~\cite{luo2021semi}     & &  & 69.44  & 57.07  &  8.32 \\
URPC~\cite{luo2022semi}  &   &   & 68.92  & 56.11  &  8.63 \\
MC-Net~\cite{wu2021semi}  & &  & 69.55  & 57.33  &  8.36 \\
MC-Net+~\cite{wu2022mutual}    &  &  & 70.07  & 57.29  &  8.50 \\
MCF~\cite{wang2023mcf} &  &   & 70.49  & 57.91  &  8.30  \\
CDMA~\cite{zhong2023semi} & & & 67.28  & 54.57  &  8.65 \\
CauSSL~\cite{miao2023caussl} & & & 71.12  & 58.69  &  8.32 \\
BS-Net~\cite{he2023bilateral} & & & 71.30  & 58.95  &  8.05 \\
BCP~\cite{bai2023bidirectional} &  & & 70.80  & 59.38  &  7.98 \\
CAML~\cite{gao2023correlation}   & &   & 69.04 & 56.16  &  8.82 \\
\hline
\textbf{MC-seg} (Ours)    &   &   & \textcolor{blue}{72.21}  & \textcolor{blue}{60.14}  &  \textcolor{blue}{7.95} \\
\textbf{KnowSAM} (Ours)   &   &  & \textcolor{red}{73.41}  & \textcolor{red}{61.50}  & \textcolor{red}{7.85}  \\
  \hline
  \end{tabular}\vspace{-0.15cm}
\end{table}

\begin{table*}[t!]
  \centering
  \scriptsize
  \renewcommand{\arraystretch}{1.1}
  \setlength\tabcolsep{3.1pt}
  \caption{\zh{Ablative results on the effects of different maps in the proposed HAM}.}
  \label{tab_abi_views}\vspace{-0.15cm}
\begin{tabular}{c c|c c c |c| c c c |c| c c c |c| c c c |c| c c c}
\hline
 \multicolumn{2}{c|}{Scans used} & \multicolumn{3}{c|}{Views} & \multirow{2}{*}{} & \multicolumn{3}{c|}{Metrics} & \multirow{2}{*}{} & \multicolumn{3}{c|}{Metrics} & \multirow{2}{*}{} & \multicolumn{3}{c|}{Metrics} & \multirow{2}{*}{} & \multicolumn{3}{c}{Metrics}\\
\cline{1-5}\cline{7-9}\cline{11-13}\cline{15-17}\cline{19-21}
        Labeled & Unlabeled    & Pred.    & $\mathcal H$ & $\mathcal M$ &   & Dice (\%)     & IoU (\%)    & HD95  &   & Dice (\%)     & IoU (\%)    & HD95 &   & Dice (\%)     & IoU (\%)    & HD95  &   & Dice (\%)     & IoU (\%)    & HD95 \\
\hline

\multirow{3}{*}{10\%} & \multirow{3}{*}{90\%} &  \checkmark &        &       &\multirow{6}{*}{\begin{sideways}Kvasir\end{sideways}}  &83.01 &75.17 &4.71  &\multirow{6}{*}{\begin{sideways}ISIC-2018\end{sideways}}  &83.99 &75.42 &5.19   &\multirow{6}{*}{\begin{sideways}TN3K\end{sideways}}  &76.53 &67.09 &4.86  &\multirow{6}{*}{\begin{sideways}ACDC\end{sideways}}  &88.28 &79.67 &4.35\\
& & \checkmark    &    \checkmark    &                & &83.74 &76.67 &4.58   &  &84.36 &76.15 &5.03  & &77.29 &68.22 &4.77   &  &88.85 &80.54 &1.35\\
& & \checkmark   &   \checkmark       &   \checkmark     &  &84.47 &77.00 &4.69   &  &84.44 &76.36 &5.09  &  &78.57 &69.42 &4.88   &  &89.08 &81.00 &1.35\\

\cline{1-2}\cline{3-5}\cline{7-9}\cline{11-13}\cline{15-17}\cline{19-21}
\multirow{3}{*}{30\%} & \multirow{3}{*}{70\%} &    \checkmark     &        &       &   &86.58 &79.69 &4.51     &  &83.94 &75.67 &5.17  &   &78.68 &69.71 &4.83     &  &88.63 &80.37 &5.14   \\
&        &    \checkmark    &   \checkmark       &        &   &87.08 &80.53 &4.29   &  &84.29 &76.37 &5.12   &   &80.12 &71.22 &4.71   &  &89.18 &81.16 &4.06\\
&        &     \checkmark   &    \checkmark      &  \checkmark      & &87.20 &80.53 &4.47  & &86.06 &77.84 &5.04 & &80.18 &71.50 &4.65  & &90.22 &82.76 &1.21\\
\hline

\end{tabular}\vspace{-0.25cm}
\end{table*}

\begin{table*}[t!]
  \centering
  \scriptsize
  \renewcommand{\arraystretch}{1.1}
  \setlength\tabcolsep{2.6pt}
  \caption{\zh{Ablation Study on the effectiveness of the proposed LPS.}}
  \label{tab_box_Strategy}\vspace{-0.15cm}
\begin{tabular}{c c|c|c|c| c c c |c| c c c |c| c c c |c| c c c}
\hline
 \multicolumn{2}{c|}{Scans used}  & \multirow{2}*{\textbf{Methods}} & \multirow{2}*{\textbf{Strategy}} &  & \multicolumn{3}{c|}{Metrics} &  & \multicolumn{3}{c|}{Metrics} &  & \multicolumn{3}{c|}{Metrics} &  & \multicolumn{3}{c}{Metrics}\\
\cline{1-2}\cline{6-8}\cline{10-12}\cline{14-16}\cline{18-20}
Labeled & Unlabeled &  &  &  & Dice (\%)     & IoU (\%)    & HD95  &   & Dice (\%)     & IoU (\%)    & HD95 &   & Dice (\%)     & IoU (\%)    & HD95 &   & Dice (\%)     & IoU (\%)    & HD95\\
\hline

\multirow{4}{*}{10\%} & \multirow{4}{*}{90\%} & \multirow{2}{*}{SAM} &  Box &\multirow{8}{*}{\begin{sideways}Kvasir\end{sideways}}  &86.60 &80.24 &4.37 &\multirow{8}{*}{\begin{sideways}ISIC-2018\end{sideways}} &87.59 &79.61 &4.91 &\multirow{8}{*}{\begin{sideways}TN3K\end{sideways}}&81.84 &72.73 &4.46     &\multirow{8}{*}{\begin{sideways}ACDC\end{sideways}} &83.24 &72.19 &4.93\\
&   & &   LPS     &  &87.29 &81.14 &4.24   &  &87.90 &80.26 &4.83  &  &82.63 &73.79 &4.43   &  &84.34 &73.99 &1.84\\
\cline{3-4}
&   & \multirow{2}{*}{KnowSAM} &   Box   & &86.32 &79.30 & 4.49   &  &86.70 &78.70 &4.94  &  &79.72 &70.55 &4.55   &  &88.42 &80.00 &4.27\\
&   & &   LPS   &  &85.98 &79.25 &4.41   &  &86.51 &78.49 &4.86 &  &81.19 &72.27 &4.52   &  &89.56 &81.66 &1.28\\
\cline{1-4}\cline{6-8}\cline{10-12}\cline{14-16}\cline{18-20}
\multirow{4}{*}{30\%} & \multirow{4}{*}{70\%} & \multirow{2}{*}{SAM} & Box    & &86.94 &80.71 &4.33  & &87.54 &79.38 &4.77  & &82.71 &73.74 &4.42  & &84.25 &73.55 &3.72\\
&    &  &  LPS  & &90.01 &84.78 &3.98  & &88.08 &80.45 &4.82 & &82.75 &73.75 &4.48  & &85.82 &75.79 &1.29\\
\cline{3-4}
&    & \multirow{2}{*}{KnowSAM} &  Box   & &85.22 &78.40 &4.43  & &87.29 &79.27 &4.79 & &81.11 &72.49 &4.55  & &89.42 &81.47 &3.64\\
&    &  &  LPS   &  &88.74 &82.93 &4.22   &  &87.22 &79.72 &4.81 &  &81.21 &72.18 &4.67   &  &91.13 &84.11 &1.29\\
\hline

\end{tabular}\vspace{-0.25cm}
\end{table*}

\begin{table*}[t!]
  \centering
  \scriptsize
  \renewcommand{\arraystretch}{1.1}
  \setlength\tabcolsep{3.4pt}
  \caption{\zh{Ablation study on the effectiveness of the proposed UGDA strategy}.}
  \label{tab_DA_Strategy}\vspace{-0.15cm}
\begin{tabular}{c c|c|c| c c c |c| c c c |c| c c c |c| c c c}
\hline
 \multicolumn{2}{c|}{Scans used}  & \multirow{2}*{\textbf{Strategy}} &  & \multicolumn{3}{c|}{Metrics} &  & \multicolumn{3}{c|}{Metrics} &  & \multicolumn{3}{c|}{Metrics} &  & \multicolumn{3}{c}{Metrics}\\
\cline{1-2}\cline{5-7}\cline{9-11}\cline{13-15}\cline{17-19}
        Labeled & Unlabeled &  &   & Dice (\%)     & IoU (\%)    & HD95  &   & Dice (\%)     & IoU (\%)    & HD95 &   & Dice (\%)     & IoU (\%)    & HD95 &   & Dice (\%)     & IoU (\%)    & HD95\\
\hline

\multirow{2}{*}{10\%} & \multirow{2}{*}{90\%} &  w/o UGDA  &\multirow{4}{*}{\begin{sideways}Kvasir\end{sideways}}  &80.45 &71.81 &5.04 &\multirow{4}{*}{\begin{sideways}ISIC-2018\end{sideways}} &83.47 &75.04 &5.20 &\multirow{4}{*}{\begin{sideways}TN3K\end{sideways}}  &76.94 &67.05 &5.04 &\multirow{4}{*}{\begin{sideways}ACDC\end{sideways}} &85.26 &75.45 &4.74 \\
&    &   w/ UGDA   &  &84.47 &77.00 &4.69   &  &84.44 &76.36 &5.09 &  &78.57 &69.42 &4.88   &  &89.08 &81.00 &1.35\\
\cline{1-3}\cline{5-7}\cline{9-11}\cline{13-15}\cline{17-19}
\multirow{2}{*}{30\%} & \multirow{2}{*}{70\%} & w/o UGDA   & &85.84 &78.44  & 4.65 &  &84.45 &75.75 &5.03 & &77.95 &68.59 &4.78  & &89.42 &81.49 &4.13\\
&     &  w/ UGDA  & &87.20 &80.53 &4.47  & &86.06 &77.84 &5.04 & &80.18 &71.50 &4.65  & &90.22 &82.76 &1.21 \\
\hline

\end{tabular}\vspace{-0.25cm}
\end{table*}

\begin{table*}[t!]
  \centering
  \scriptsize
  \renewcommand{\arraystretch}{1.1}
  \setlength\tabcolsep{2.2pt}
  \caption{\zh{Ablation study on the effects of different loss functions}.}
  \label{tab_abi_loss}\vspace{-0.15cm}
\begin{tabular}{c c|c c c c |c| c c c |c| c c c |c| c c c |c| c c c}
\hline
 \multicolumn{2}{c|}{Scans used} & \multicolumn{4}{c|}{Losses} & \multirow{2}{*}{} & \multicolumn{3}{c|}{Metrics} & \multirow{2}{*}{} & \multicolumn{3}{c|}{Metrics} & \multirow{2}{*}{} & \multicolumn{3}{c|}{Metrics} & \multirow{2}{*}{} & \multicolumn{3}{c}{Metrics}\\
\cline{1-6}\cline{8-10}\cline{12-14}\cline{16-18}\cline{20-22}
        Labeled & Unlabeled & $\mathcal {L}_{sup}$    & $\mathcal {L}_{ent}$    & $\mathcal {L}_{mut}$ & $\mathcal {L}_{kd}$ &   & Dice (\%)     & IoU (\%)    & HD95  &   & Dice (\%)     & IoU (\%)    & HD95 &   & Dice (\%)     & IoU (\%)    & HD95 &   & Dice (\%)     & IoU (\%)    & HD95   \\
\hline
 \multirow{6}{*}{10\%} & \multirow{6}{*}{90\%} &  \checkmark &       &       &   &\multirow{12}{*}{\begin{sideways}Kvasir\end{sideways}} &81.94 &73.94 &4.96 &\multirow{12}{*}{\begin{sideways}ISIC-2018\end{sideways}} &83.58 &74.95 &5.15  &\multirow{12}{*}{\begin{sideways}TN3K\end{sideways}} &77.76 &68.43 &4.85 &\multirow{12}{*}{\begin{sideways}ACDC\end{sideways}}  &86.70 &77.34 &6.58\\
&        &    \checkmark    &    \checkmark      &       &    & &83.56 &75.82 &4.84  & &84.10 &76.07 &5.13    & &76.82 &67.67 &4.84 &  &88.61 &80.16 &2.53  \\
&        &    \checkmark    &          &  \checkmark      &       & &83.40 &76.30 &4.71  &  &84.60 &76.54 &5.08   & &77.55 &68.32&4.81 &  &87.96 &79.26 &4.38\\
&        &    \checkmark    &          &        & \checkmark      & &85.86 &79.23 &4.27  &  &86.49 &78.70 &4.87   & &79.28 &69.98 &4.59 &  &88.39 &79.88 &2.97\\ 
&        &     \checkmark   &   \checkmark       &   \checkmark  &    & &84.47 &77.00 &4.69 &  &84.44 &76.36 &5.09   &  &78.57 &69.42 &4.88 &  &89.08 &81.00 &1.35\\
&        &     \checkmark   &   \checkmark       &   \checkmark     & \checkmark & &85.98 &79.25 &4.41 & &86.51 &78.49 &4.86   & &81.19 &72.27 &4.52   &  &89.56 &81.66 &1.28  \\
\cline{1-6}\cline{8-10}\cline{12-14}\cline{16-18}\cline{20-22}
\multirow{6}{*}{30\%} & \multirow{6}{*}{70\%} &    \checkmark     &        &        &   & &84.60 &77.66 &4.75 & &86.15 &78.05 &4.95    & &80.08 &71.31 &4.66   &  &89.34 &81.40 &4.18\\
&        &    \checkmark    &   \checkmark       &        &   & &87.17 &80.42 &4.36  & &86.48 &78.37 &4.95   & &79.59 &70.92 &4.68  &  &89.68 &81.98 &4.57\\
&        &    \checkmark    &          &  \checkmark      &     & &87.38 &81.03 &4.38  &  &86.01 &77.95 &5.00  & &79.59 &70.76 &4.71  &  &89.62 &81.77 &3.81\\
&        &    \checkmark    &          &        & \checkmark    & &88.37 &82.04 &4.23  &  &86.41 &78.39 &4.91  & &81.00 &72.29 &4.51  &  &90.07 &82.44 &3.39\\ 
&        &     \checkmark   &    \checkmark      &  \checkmark      &  & &87.20 &80.53 &4.47  & &86.06 &77.84 &5.04   & &80.18 &71.50 &4.65  &  &90.22 &82.76 &1.21\\
&        &     \checkmark   &    \checkmark      &  \checkmark      &   \checkmark  & &88.74 &82.93 &4.22 & &87.22 &79.72 &4.81   & &81.21 &72.18 &4.67  &  &91.13 &84.11 &1.29\\
\hline
\end{tabular}\vspace{-0.05cm}
\end{table*}

\subsubsection{Effectiveness of Learnable Prompt Strategy}

In the context of SAM's prompting strategy, we propose the LPS to mitigate the substantial performance degradation caused by the use of specific points or box coordinates derived from erroneous predictions as prompts. To evaluate the effectiveness of LPS, we conduct ablation experiments, comparing LPS with a traditional strategy that employs specific bounding box coordinates (denoted as ``Box"). As illustrated in Table \ref{tab_box_Strategy}, we report the predicted results of SAM and the fused predictions $\hat{Y}_f$ (\ie, the outputs of KnowSAM). \zh{It is observed that our LPS outperforms the approach utilizing the ``Box" strategy. This is attributed to a learnable feature prompt provided by the LPS, which effectively mitigates the performance degradation of SAM that can result from inaccurate hard coordinates. Consequently, this enhancement improves SAM's effectiveness in medical segmentation tasks}.

\subsubsection{Effectiveness of the Uncertainty-Guided Data Augmentation}
In this study, we introduce uncertainty information into the traditional data augmentation to guide the copy-paste process between labeled and unlabeled data. To demonstrate the effectiveness of UGDA, we conduct ablation experiments on the polyp segmentation task. As shown in Table \ref{tab_DA_Strategy}, the dice coefficient of the UGDA method increases from $80.45\%$ to $84.47\%$, and HD95 decreases from $5.04$ to $4.69$ compared to the traditional data augmentation approach, with $10\%$ labeled images on the Kvasir dataset. \zh{In UGDA, it is crucial to highlight that the labels for the mixed images are a combination of ground truths from labeled data and pseudo-labels from unlabeled data. This strategy enables the network to capture common semantics between labeled and unlabeled data, thereby boosting the generalization of our model}.

\begin{figure*}[t]
	\centering
	\begin{overpic}[width=0.98\textwidth]{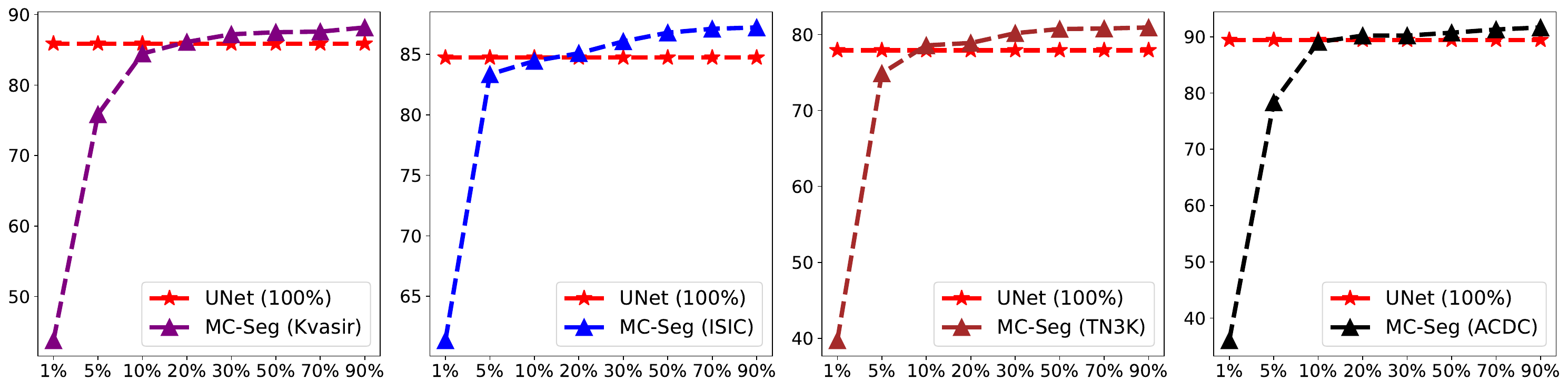}
    \end{overpic}\vspace{-0.25cm}
	\caption{\zh{Dice performance of our MC-Seg under different label ratios on the four distinct medical segmentation datasets.} 
	}
    \label{fig_ratios_four}\vspace{-0.25cm}
\end{figure*}

\begin{table*}[t!]
  \centering
  \scriptsize
  \renewcommand{\arraystretch}{1.1}
  \setlength\tabcolsep{3.0pt}
  \caption{\zh{Quantitative results of different SAM-based semi-supervised methods ($^{\dagger}$ denotes that the outputs of SAM are taken as segmentation results)}.}
  \label{tab:tab_abi_samSSL}\vspace{-0.15cm}
\begin{tabular}{c c|r|c| c c c |c| c c c |c| c c c |c| c c c}
\hline
 \multicolumn{2}{c|}{Scans used}  & \multirow{2}*{\textbf{Methods}} &  & \multicolumn{3}{c|}{Metrics} &  & \multicolumn{3}{c|}{Metrics} &  & \multicolumn{3}{c|}{Metrics} &  & \multicolumn{3}{c}{Metrics} \\
\cline{1-2}\cline{5-7}\cline{9-11}\cline{13-15}\cline{17-19}
        Labeled & Unlabeled &  &   & Dice (\%)     & IoU (\%)    & HD95  &   & Dice (\%)     & IoU (\%)    & HD95 &   & Dice (\%)     & IoU (\%)    & HD95 &   & Dice (\%)     & IoU (\%)    & HD95\\

\hline
\multirow{3}{*}{10\%} & \multirow{3}{*}{90\%} &  SemiSAM~\cite{zhang2023semisam} &\multirow{6}{*}{\begin{sideways}Kvasir\end{sideways}}  &69.94 &59.84 &5.39 &\multirow{6}{*}{\begin{sideways}ISIC-2018\end{sideways}} &79.93 &71.11 &5.27 &\multirow{6}{*}{\begin{sideways}TN3K\end{sideways}}  &66.58 &56.03 &4.82 &\multirow{6}{*}{\begin{sideways}ACDC\end{sideways}} &82.70 &72.17 &2.85\\
&    &  CPC-SAM$^{\dagger}$~\cite{miao2024cross}    &  &82.24 &75.01 &4.43   &  &87.96 &80.37 &4.54 &  &75.02 &64.87 &4.49  &  &85.15 &75.08 &3.97\\
&    &   KnowSAM (Ours)  &  &85.98 &79.25 &4.41   &  &86.51 &78.49 &4.86  &  &81.19 &72.27 &4.52  &  &89.56 &81.66 &1.28\\
\cline{1-3}\cline{5-7}\cline{9-11}\cline{13-15}\cline{17-19}

\multirow{3}{*}{30\%} & \multirow{3}{*}{70\%} & SemiSAM~\cite{zhang2023semisam}    & &81.80 &73.70 &4.54  & &82.62 &73.96 &4.85  & &72.29 &61.85 &4.64  & &86.89 &77.83 &1.98\\
&    &   CPC-SAM$^{\dagger}$~\cite{miao2024cross}     &  &87.36 &81.05 &4.07   &  &88.11 &80.33 &4.50   &  &77.73 &67.83 &4.39   &  &88.61 &80.08 &1.24\\
&    &   KnowSAM (Ours)    &  &88.74 &82.93 &4.22   &  &87.22 &79.72 &4.81   &  &81.21 &72.18 &4.67  &  &91.13 &84.11 &1.29\\
\hline

\end{tabular}\vspace{-0.25cm}
\end{table*}

\begin{table*}[t!]
  \centering
  \scriptsize
  \renewcommand{\arraystretch}{1.1}
  \setlength\tabcolsep{3.1pt}
  \caption{\zh{Comparison results of our method using different SAM-based foundation models}.}
  \label{tab:tab_abi_medsam}\vspace{-0.05cm}
\begin{tabular}{c c|r|c| c c c |c| c c c |c| c c c |c| c c c}
\hline
 \multicolumn{2}{c|}{Scans used}  & \multirow{2}*{\textbf{SAM-V}} &  & \multicolumn{3}{c|}{Metrics} &  & \multicolumn{3}{c|}{Metrics} &  & \multicolumn{3}{c|}{Metrics} &  & \multicolumn{3}{c}{Metrics}\\
\cline{1-2}\cline{5-7}\cline{9-11}\cline{13-15}
        Labeled & Unlabeled &  &   & Dice (\%)     & IoU (\%)    & HD95  &   & Dice (\%)     & IoU (\%)    & HD95 &   & Dice (\%)     & IoU (\%)    & HD95 &   & Dice (\%)     & IoU (\%)    & HD95\\
\hline
\multirow{2}{*}{10\%} & \multirow{2}{*}{90\%} &  MedSAM~\cite{ma2024segment} &\multirow{4}{*}{\begin{sideways}Kvasir\end{sideways}}  &75.02 &65.79 &5.35 &\multirow{4}{*}{\begin{sideways}ISIC-2018\end{sideways}} &83.89 &76.06 &5.07  &\multirow{4}{*}{\begin{sideways}TN3K\end{sideways}}  &71.60 &61.85 &5.02 &\multirow{4}{*}{\begin{sideways}ACDC\end{sideways}} &82.73 &72.35 &2.77\\
&    &   SAM~\cite{kirillov2023segment}    &  &85.98 &79.25 &4.41   &  &86.51 &78.49 &4.86  &  &81.19 &72.27 &4.52   &  &89.56 &81.66 &1.28\\
\cline{1-3}\cline{5-7}\cline{9-11}\cline{13-15}\cline{17-19}
\multirow{2}{*}{30\%} & \multirow{2}{*}{70\%} & MedSAM~\cite{ma2024segment}     & &82.89 &75.02 &5.00  & &85.60 &77.27 &4.92  & &76.73 &67.32 &4.81  & &89.58 &81.78 &1.57\\
&     &  SAM~\cite{kirillov2023segment}    &  &88.74 &82.93 &4.22   &  &87.22 &79.72 &4.81  &  &81.21 &72.18 &4.67   &  &91.13 &84.11 &1.29\\
\hline

\end{tabular}\vspace{-0.25cm}
\end{table*}

\subsubsection{Effectiveness of Different Loss Functions}

The total loss function of our method consists of four components: supervision loss $\mathcal L_{sup}$, entropy loss $\mathcal L_{ent}$, mutual consistency loss $\mathcal L_{mut}$, and distillation loss $\mathcal L_{kd}$. The entropy loss function aims to eliminate the inherent uncertainty of the subnets. The consistency loss function is designed to reduce the cognitive differences between subnets for the same input, and it also serves as supervision for the dissimilarity map ($\mathcal{M}$). It is noteworthy that the inclusion of the $\mathcal{L}_{kd}$ loss function indicates whether SAM is used for distillation. To investigate the contribution of each loss function, we start with the supervision loss as the base and gradually add other loss functions. 
\zh{
Table~\ref{tab_abi_loss} presents the ablative results across four distinct medical segmentation tasks. The model exhibits a progressive enhancement in segmentation accuracy with the addition of each component, thereby validating the effectiveness of each key element. Notably, when only $\mathcal L_{sup}$ and $\mathcal L_{kd}$ are combined, significant performance improvements are achieved across all tasks. This underscores the pivotal role of SAM-based distillation in enhancing model performance, particularly by mitigating cognitive disparities between the two subnets and strengthening segmentation capabilities.
}

\subsection{Discussion}

\subsubsection{Model Performance with Different Label Ratios}

\zh{We further conduct experiments to investigate the impact of different label ratios on model performance. Fig.~\ref{fig_ratios_four} shows the Dice scores on the Kvasir, ISIC-2018, TN3K, and ACDC datasets. Moreover, we introduce a fully supervised model,} \zh{such as UNet, as the benchmark for comparison. As depicted in Fig.~\ref{fig_ratios_four}, the performance of our MC-Seg tends to improve with an increasing proportion of labeled data, which is anticipated since a larger number of labeled samples offer more precise guidance during training. Furthermore, our MC-Seg model achieves nearly equivalent performance to the fully supervised UNet, even with a relatively small fraction of labeled data, demonstrating its robustness and effectiveness.}

\subsubsection{Comparison with SAM-based Semi-supervised Methods}

\zh{
In this study, we harness the robust segmentation capabilities of the SAM through distillation learning to enhance the performance of semi-supervised medical image segmentation. To assess various SAM-based semi-supervised segmentation approaches, we compare the proposed model with two other SAM-based methods: SemiSAM~\cite{zhang2023semisam} and CPC-SAM~\cite{miao2024cross}. It is important to highlight that in CPC-SAM, SAM's outputs are treated as the definitive segmentation outcomes, whereas in our approach and SemiSAM, the outputs from the UNet model are regarded as the segmentation maps. Notably, UNet offers a significant reduction in inference cost due to its reduced computational demands compared to SAM. Table \ref{tab:tab_abi_samSSL} shows a quantitative comparison of SAM-based semi-supervised methods with $10\%$ and $30\% $labeled data. It can be observed that our model consistently outperforms SemiSAM across all datasets, thereby validating its effectiveness. Furthermore, our KnowSAM model delivers promising results on several datasets with a lower inference cost, underscoring its ability to boost segmentation performance.
}

\subsubsection{Effects of Our Method using Different Foundation Models}

\zh{Recently, the MedSAM~\cite{ma2024segment} has fine-tuned SAM to better adapt it for downstream tasks in the medical imaging field. To further investigate our choice of SAM, we conduct an ablation study. In this experiment, both SAM and MedSAM are trained with their original parameters frozen, with an adapter module incorporated for fine-tuning. As shown in Table \ref{tab:tab_abi_medsam}, our model, which employs SAM, consistently outperformed the one that employs MedSAM across all tasks. We believe that while MedSAM could potentially excel on the specific datasets for which it has been tailored, the fine-tuning process may diminish the robust generalization capabilities that are an integral strength of the original SAM model.}

\begin{figure}
	\centering
	\begin{overpic}[width=0.4\textwidth]{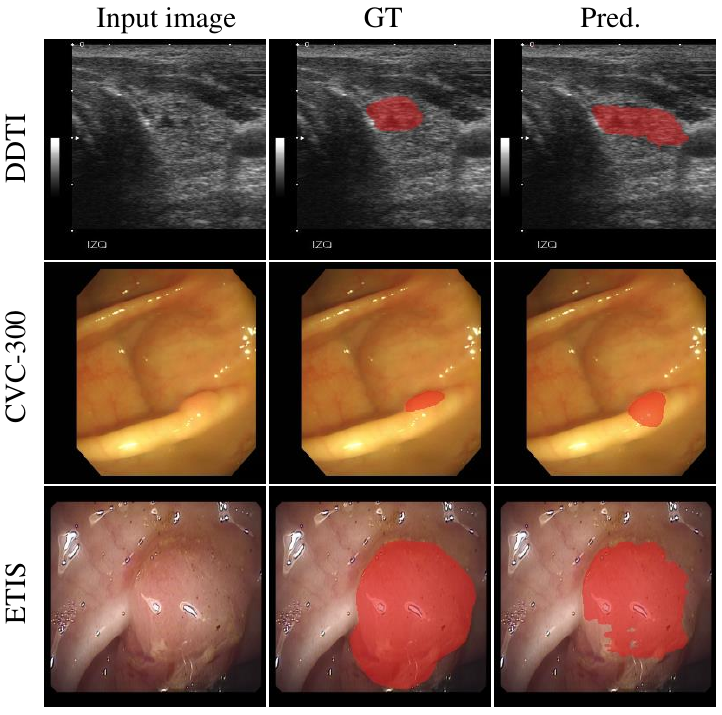}
    \end{overpic}\vspace{-0.2cm}
	\caption{\zh{Visualizations of some failure cases generated by our model.} 
	}
    \label{fig_fail}\vspace{-0.25cm}
\end{figure}


\subsubsection{Analysis of Failure Cases}


\zh{Qualitative and quantitative evaluations confirm the effectiveness and superiority of our model. However, it faces difficulties in accurately segmenting tumors, particularly in regions with ambiguity and complex backgrounds. Some failure cases generated by our model are depicted in \figref{fig_fail}. In the $1^{st}$ row, it is evident that the tumor area is heavily obscured by the background, with very indistinct boundaries, posing a significant challenge for precise tumor segmentation. It can be seen that under this circumstance, our model struggles to identify and segment the tumor. In the $2^{nd}$ and $3^{rd}$ rows, our model is capable of locating the primary regions of polyps but generates some imprecise fragments in two samples from unseen datasets. The data distribution between the training set and unseen test} \zh{sets exhibits a significant difference, impeding the model's ability to generalize effectively. Consequently, dealing with ambiguous areas in semi-supervised segmentation tasks is desired to be investigated in future work. Moreover, exploring adaptive methods to enhance segmentation accuracy on unseen datasets is also a focus for future investigation}.

\section{Conclusion}

In this paper, we have proposed a novel KnowSAM framework for semi-supervised medical image segmentation. By employing the multi-view co-training paradigm, our model enables two subnets to learn from SAM through knowledge distillation, enhancing the generalization capabilities of the subnets and mitigating the impact of noise in unlabeled data. Additionally, we present the UGDA strategy to enhance the model's robustness through the interaction between labeled and unlabeled data. This effective strategy also avoids incurring additional costs during inference. Experimental results across various datasets validate the superiority of the proposed KnowSAM. It is worth noting that our method can be easily extended to enhance other semi-supervised segmentation tasks.

{
\bibliographystyle{IEEEtran}
\bibliography{tmi}

\begin{thebibliography}{10}
\providecommand{\url}[1]{#1}
\csname url@samestyle\endcsname
\providecommand{\newblock}{\relax}
\providecommand{\bibinfo}[2]{#2}
\providecommand{\BIBentrySTDinterwordspacing}{\spaceskip=0pt\relax}
\providecommand{\BIBentryALTinterwordstretchfactor}{4}
\providecommand{\BIBentryALTinterwordspacing}{\spaceskip=\fontdimen2\font plus
\BIBentryALTinterwordstretchfactor\fontdimen3\font minus \fontdimen4\font\relax}
\providecommand{\BIBforeignlanguage}[2]{{%
\expandafter\ifx\csname l@#1\endcsname\relax
\typeout{** WARNING: IEEEtran.bst: No hyphenation pattern has been}%
\typeout{** loaded for the language `#1'. Using the pattern for}%
\typeout{** the default language instead.}%
\else
\language=\csname l@#1\endcsname
\fi
#2}}
\providecommand{\BIBdecl}{\relax}
\BIBdecl

\bibitem{zhang2017deep}
Y.~Zhang, L.~Yang, J.~Chen, M.~Fredericksen, D.~P. Hughes, and D.~Z. Chen, ``Deep adversarial networks for biomedical image segmentation utilizing unannotated images,'' in \emph{Proc. MICCAI}.\hskip 1em plus 0.5em minus 0.4em\relax Springer, 2017, pp. 408--416.

\bibitem{zhou2024uncertainty}
T.~Zhou, Y.~Zhou, G.~Li, G.~Chen, and J.~Shen, ``Uncertainty-aware hierarchical aggregation network for medical image segmentation,'' \emph{IEEE TCSVT}, vol.~34, no.~8, pp. 7440--7453, 2024.

\bibitem{tajbakhsh2020embracing}
N.~Tajbakhsh, L.~Jeyaseelan, Q.~Li, J.~N. Chiang, Z.~Wu, and X.~Ding, ``Embracing imperfect datasets: A review of deep learning solutions for medical image segmentation,'' \emph{Medical Image Analysis}, vol.~63, p. 101693, 2020.

\bibitem{bai2023bidirectional}
Y.~Bai, D.~Chen, Q.~Li, W.~Shen, and Y.~Wang, ``Bidirectional copy-paste for semi-supervised medical image segmentation,'' in \emph{Proc. CVPR}, 2023, pp. 11\,514--11\,524.

\bibitem{gao2023correlation}
S.~Gao, Z.~Zhang, J.~Ma, Z.~Li, and S.~Zhang, ``Correlation-aware mutual learning for semi-supervised medical image segmentation,'' in \emph{Proc. MICCAI}.\hskip 1em plus 0.5em minus 0.4em\relax Springer, 2023, pp. 98--108.

\bibitem{luo2021semi}
X.~Luo, J.~Chen, T.~Song, and G.~Wang, ``Semi-supervised medical image segmentation through dual-task consistency,'' in \emph{Proc. AAAI}, vol.~35, no.~10, 2021, pp. 8801--8809.

\bibitem{luo2022semi}
X.~Luo, G.~Wang, W.~Liao, J.~Chen, T.~Song, Y.~Chen, S.~Zhang, D.~N. Metaxas, and S.~Zhang, ``Semi-supervised medical image segmentation via uncertainty rectified pyramid consistency,'' \emph{Medical Image Analysis}, vol.~80, p. 102517, 2022.

\bibitem{wu2021semi}
Y.~Wu, M.~Xu, Z.~Ge, J.~Cai, and L.~Zhang, ``Semi-supervised left atrium segmentation with mutual consistency training,'' in \emph{Proc. MICCAI}.\hskip 1em plus 0.5em minus 0.4em\relax Springer, 2021, pp. 297--306.

\bibitem{wang2023mcf}
Y.~Wang, B.~Xiao, X.~Bi, W.~Li, and X.~Gao, ``Mcf: Mutual correction framework for semi-supervised medical image segmentation,'' in \emph{Proc. CVPR}, 2023, pp. 15\,651--15\,660.

\bibitem{tarvainen2017mean}
A.~Tarvainen and H.~Valpola, ``Mean teachers are better role models: Weight-averaged consistency targets improve semi-supervised deep learning results,'' \emph{Proc. NeurIPS}, vol.~30, 2017.

\bibitem{xu2023ambiguity}
Z.~Xu, Y.~Wang, D.~Lu, X.~Luo, J.~Yan, Y.~Zheng, and R.~K.-y. Tong, ``Ambiguity-selective consistency regularization for mean-teacher semi-supervised medical image segmentation,'' \emph{Medical Image Analysis}, vol.~88, p. 102880, 2023.

\bibitem{yu2019uncertainty}
L.~Yu, S.~Wang, X.~Li, C.-W. Fu, and P.-A. Heng, ``Uncertainty-aware self-ensembling model for semi-supervised 3{D} left atrium segmentation,'' in \emph{Proc. MICCAI}.\hskip 1em plus 0.5em minus 0.4em\relax Springer, 2019, pp. 605--613.

\bibitem{wang2022semi}
K.~Wang, B.~Zhan, C.~Zu, X.~Wu, J.~Zhou, L.~Zhou, and Y.~Wang, ``Semi-supervised medical image segmentation via a tripled-uncertainty guided mean teacher model with contrastive learning,'' \emph{Medical Image Analysis}, vol.~79, p. 102447, 2022.

\bibitem{shen2023co}
Z.~Shen, P.~Cao, H.~Yang, X.~Liu, J.~Yang, and O.~R. Zaiane, ``Co-training with high-confidence pseudo labels for semi-supervised medical image segmentation,'' \emph{arXiv preprint arXiv:2301.04465}, 2023.

\bibitem{hu2021semi}
X.~Hu, D.~Zeng, X.~Xu, and Y.~Shi, ``Semi-supervised contrastive learning for label-efficient medical image segmentation,'' in \emph{Proc. MICCAI}.\hskip 1em plus 0.5em minus 0.4em\relax Springer, 2021, pp. 481--490.

\bibitem{he2020dense}
Y.~He, G.~Yang, J.~Yang, Y.~Chen, Y.~Kong, J.~Wu \emph{et~al.}, ``Dense biased networks with deep priori anatomy and hard region adaptation: Semi-supervised learning for fine renal artery segmentation,'' \emph{Medical Image Analysis}, vol.~63, p. 101722, 2020.

\bibitem{kirillov2023segment}
A.~Kirillov, E.~Mintun, N.~Ravi, H.~Mao, C.~Rolland, L.~Gustafson, T.~Xiao, S.~Whitehead, A.~C. Berg, W.-Y. Lo \emph{et~al.}, ``Segment anything,'' \emph{arXiv preprint arXiv:2304.02643}, 2023.

\bibitem{chen2023sam}
T.~Chen, L.~Zhu, C.~Ding, R.~Cao, S.~Zhang, Y.~Wang \emph{et~al.}, ``{SAM} fails to segment anything?--sam-adapter: Adapting sam in underperformed scenes: Camouflage, shadow, and more,'' \emph{arXiv preprint arXiv:2304.09148}, 2023.

\bibitem{wu2023medical}
J.~Wu, R.~Fu, H.~Fang, Y.~Liu, Z.~Wang, Y.~Xu, Y.~Jin, and T.~Arbel, ``Medical sam adapter: Adapting segment anything model for medical image segmentation,'' \emph{arXiv preprint arXiv:2304.12620}, 2023.

\bibitem{li2024concatenate}
S.~Li, L.~Qi, Q.~Yu, J.~Huo, Y.~Shi, and Y.~Gao, ``Concatenate, fine-tuning, re-training: A {SAM}-enabled framework for semi-supervised 3d medical image segmentation,'' \emph{arXiv preprint arXiv:2403.11229}, 2024.

\bibitem{xu2024esp}
Q.~Xu, J.~Li, X.~He, Z.~Liu, Z.~Chen, W.~Duan \emph{et~al.}, ``Esp-medsam: Efficient self-prompting sam for universal domain-generalized medical image segmentation,'' \emph{arXiv preprint arXiv:2407.14153}, 2024.

\bibitem{zhang2023semisam}
Y.~Zhang, Y.~Cheng, and Y.~Qi, ``Semisam: Exploring sam for enhancing semi-supervised medical image segmentation with extremely limited annotations,'' \emph{arXiv preprint arXiv:2312.06316}, 2023.

\bibitem{zhang2023segment}
Y.~Zhang, S.~Hu, C.~Jiang, Y.~Cheng, and Y.~Qi, ``Segment anything model with uncertainty rectification for auto-prompting medical image segmentation,'' \emph{arXiv preprint arXiv:2311.10529}, 2023.

\bibitem{zeng2023ss}
L.-L. Zeng, K.~Gao, D.~Hu, Z.~Feng, C.~Hou, P.~Rong, and W.~Wang, ``{SS-TBN}: A semi-supervised tri-branch network for covid-19 screening and lesion segmentation,'' \emph{IEEE TPAMI}, vol.~45, no.~8, pp. 10\,427--10\,442, 2023.

\bibitem{yao2022enhancing}
H.~Yao, X.~Hu, and X.~Li, ``Enhancing pseudo label quality for semi-supervised domain-generalized medical image segmentation,'' in \emph{Proc. AAAI}, vol.~36, no.~3, 2022, pp. 3099--3107.

\bibitem{jiao2023learning}
R.~Jiao, Y.~Zhang, L.~Ding, B.~Xue, J.~Zhang, R.~Cai, and C.~Jin, ``Learning with limited annotations: a survey on deep semi-supervised learning for medical image segmentation,'' \emph{CIBM}, p. 107840, 2023.

\bibitem{wang2022ssa}
X.~Wang, Y.~Yuan, D.~Guo, X.~Huang, Y.~Cui, M.~Xia, Z.~Wang, C.~Bai, and S.~Chen, ``{SSA-N}et: Spatial self-attention network for {COVID}-19 pneumonia infection segmentation with semi-supervised few-shot learning,'' \emph{Medical Image Analysis}, vol.~79, p. 102459, 2022.

\bibitem{huang2022semi}
W.~Huang, C.~Chen, Z.~Xiong, Y.~Zhang, X.~Chen, X.~Sun, and F.~Wu, ``Semi-supervised neuron segmentation via reinforced consistency learning,'' \emph{IEEE TMI}, vol.~41, no.~11, pp. 3016--3028, 2022.

\bibitem{zheng2022double}
K.~Zheng, J.~Xu, and J.~Wei, ``Double noise mean teacher self-ensembling model for semi-supervised tumor segmentation,'' in \emph{Proc. ICASSP}.\hskip 1em plus 0.5em minus 0.4em\relax IEEE, 2022, pp. 1446--1450.

\bibitem{yang2023revisiting}
L.~Yang, L.~Qi, L.~Feng, W.~Zhang, and Y.~Shi, ``Revisiting weak-to-strong consistency in semi-supervised semantic segmentation,'' in \emph{Proc. CVPR}, 2023, pp. 7236--7246.

\bibitem{chen2021mtans}
G.~Chen, J.~Ru, Y.~Zhou, I.~Rekik, Z.~Pan, X.~Liu, Y.~Lin, B.~Lu, and J.~Shi, ``{MTANS}: multi-scale mean teacher combined adversarial network with shape-aware embedding for semi-supervised brain lesion segmentation,'' \emph{NeuroImage}, vol. 244, p. 118568, 2021.

\bibitem{peiris2023uncertainty}
H.~Peiris, M.~Hayat, Z.~Chen, G.~Egan, and M.~Harandi, ``Uncertainty-guided dual-views for semi-supervised volumetric medical image segmentation,'' \emph{Nature Machine Intelligence}, vol.~5, no.~7, pp. 724--738, 2023.

\bibitem{wu2021collaborative}
H.~Wu, G.~Chen, Z.~Wen, and J.~Qin, ``Collaborative and adversarial learning of focused and dispersive representations for semi-supervised polyp segmentation,'' in \emph{Proc. ICCV}, 2021, pp. 3489--3498.

\bibitem{xie2023deep}
Y.~Xie, Y.~Yin, Q.~Li, and Y.~Wang, ``Deep mutual distillation for semi-supervised medical image segmentation,'' in \emph{Proc. MICCAI}.\hskip 1em plus 0.5em minus 0.4em\relax Springer, 2023, pp. 540--550.

\bibitem{you2022simcvd}
C.~You, Y.~Zhou, R.~Zhao, L.~Staib, and J.~S. Duncan, ``Simcvd: Simple contrastive voxel-wise representation distillation for semi-supervised medical image segmentation,'' \emph{IEEE TMI}, vol.~41, no.~9, pp. 2228--2237, 2022.

\bibitem{you2023bootstrapping}
C.~You, W.~Dai, Y.~Min, L.~Staib, and J.~S. Duncan, ``Bootstrapping semi-supervised medical image segmentation with anatomical-aware contrastive distillation,'' in \emph{Proc. IPMI}.\hskip 1em plus 0.5em minus 0.4em\relax Springer, 2023, pp. 641--653.

\bibitem{shu2022cross}
Y.~Shu, H.~Li, B.~Xiao, X.~Bi, and W.~Li, ``Cross-mix monitoring for medical image segmentation with limited supervision,'' \emph{IEEE TMM}, vol.~25, pp. 1700--1712, 2022.

\bibitem{wang2023s}
A.~Wang, M.~Xu, Y.~Zhang, M.~Islam, and H.~Ren, ``S2me: Spatial-spectral mutual teaching and ensemble learning for scribble-supervised polyp segmentation,'' \emph{arXiv preprint arXiv:2306.00451}, 2023.

\bibitem{chen2021semi}
X.~Chen, Y.~Yuan, G.~Zeng, and J.~Wang, ``Semi-supervised semantic segmentation with cross pseudo supervision,'' in \emph{Proc. CVPR}, 2021, pp. 2613--2622.

\bibitem{zhong2023semi}
L.~Zhong, X.~Liao, S.~Zhang, and G.~Wang, ``Semi-supervised pathological image segmentation via cross distillation of multiple attentions,'' \emph{arXiv preprint arXiv:2305.18830}, 2023.

\bibitem{hinton2015distilling}
G.~Hinton, O.~Vinyals, and J.~Dean, ``Distilling the knowledge in a neural network,'' \emph{arXiv preprint arXiv:1503.02531}, 2015.

\bibitem{yun2019cutmix}
S.~Yun, D.~Han, S.~J. Oh, S.~Chun, J.~Choe, and Y.~Yoo, ``Cutmix: Regularization strategy to train strong classifiers with localizable features,'' in \emph{Proc. ICCV}, 2019, pp. 6023--6032.

\bibitem{ma2024constructing}
Q.~Ma, J.~Zhang, L.~Qi, Q.~Yu, Y.~Shi, and Y.~Gao, ``Constructing and exploring intermediate domains in mixed domain semi-supervised medical image segmentation,'' in \emph{Proc. CVPR}, 2024, pp. 11\,642--11\,651.

\bibitem{wu2022mutual}
Y.~Wu, Z.~Ge, D.~Zhang, M.~Xu, L.~Zhang, Y.~Xia, and J.~Cai, ``Mutual consistency learning for semi-supervised medical image segmentation,'' \emph{Medical Image Analysis}, vol.~81, p. 102530, 2022.

\bibitem{miao2023caussl}
J.~Miao, C.~Chen, F.~Liu, H.~Wei, and P.-A. Heng, ``Caussl: Causality-inspired semi-supervised learning for medical image segmentation,'' in \emph{Proc. ICCV}, 2023, pp. 21\,426--21\,437.

\bibitem{he2023bilateral}
A.~He, T.~Li, J.~Yan, K.~Wang, and H.~Fu, ``Bilateral supervision network for semi-supervised medical image segmentation,'' \emph{IEEE TMI}, vol.~43, no.~5, pp. 1715--1726, 2024.

\bibitem{silva2014toward}
J.~Silva, A.~Histace, O.~Romain, X.~Dray, and B.~Granado, ``Toward embedded detection of polyps in wce images for early diagnosis of colorectal cancer,'' \emph{IJCARS}, vol.~9, pp. 283--293, 2014.

\bibitem{bernal2015wm}
J.~Bernal, F.~J. S{\'a}nchez, G.~Fern{\'a}ndez-Esparrach, D.~Gil, C.~Rodr{\'\i}guez, and F.~Vilari{\~n}o, ``{WM-DOVA} maps for accurate polyp highlighting in colonoscopy: Validation vs. saliency maps from physicians,'' \emph{CMIG}, vol.~43, pp. 99--111, 2015.

\bibitem{tajbakhsh2015automated}
N.~Tajbakhsh, S.~R. Gurudu, and J.~Liang, ``Automated polyp detection in colonoscopy videos using shape and context information,'' \emph{IEEE TMI}, vol.~35, no.~2, pp. 630--644, 2015.

\bibitem{vazquez2017benchmark}
D.~V{\'a}zquez, J.~Bernal, F.~J. S{\'a}nchez, G.~Fern{\'a}ndez-Esparrach, A.~M. L{\'o}pez, A.~Romero, M.~Drozdzal, A.~Courville \emph{et~al.}, ``A benchmark for endoluminal scene segmentation of colonoscopy images,'' \emph{Journal of Healthcare Engineering}, vol. 2017, 2017.

\bibitem{jha2020kvasir}
D.~Jha, P.~H. Smedsrud, M.~A. Riegler, P.~Halvorsen, T.~de~Lange, D.~Johansen, and H.~D. Johansen, ``Kvasir-seg: A segmented polyp dataset,'' in \emph{MMM}.\hskip 1em plus 0.5em minus 0.4em\relax Springer, 2020, pp. 451--462.

\bibitem{fan2020pranet}
D.-P. Fan, G.-P. Ji, T.~Zhou, G.~Chen, H.~Fu, J.~Shen, and L.~Shao, ``Pranet: Parallel reverse attention network for polyp segmentation,'' in \emph{Proc. MICCAI}.\hskip 1em plus 0.5em minus 0.4em\relax Springer, 2020, pp. 263--273.

\bibitem{gong2023thyroid}
H.~Gong, J.~Chen, G.~Chen, H.~Li, G.~Li, and F.~Chen, ``Thyroid region prior guided attention for ultrasound segmentation of thyroid nodules,'' \emph{CIBM}, vol. 155, p. 106389, 2023.

\bibitem{wunderling2017comparison}
T.~Wunderling, B.~Golla, P.~Poudel, C.~Arens, M.~Friebe, and C.~Hansen, ``Comparison of thyroid segmentation techniques for 3{D} ultrasound,'' in \emph{Medical Imaging: Image Processing}, vol. 10133, 2017, pp. 346--352.

\bibitem{pedraza2015open}
L.~Pedraza, C.~Vargas, F.~Narv{\'a}ez, O.~Dur{\'a}n, E.~Mu{\~n}oz, and E.~Romero, ``An open access thyroid ultrasound image database,'' in \emph{Proc. ISMIPA}, vol. 9287.\hskip 1em plus 0.5em minus 0.4em\relax SPIE, 2015, pp. 188--193.

\bibitem{codella2019skin}
N.~Codella, V.~Rotemberg, P.~Tschandl, M.~E. Celebi, S.~Dusza, D.~Gutman \emph{et~al.}, ``Skin lesion analysis toward melanoma detection 2018: A challenge hosted by the international skin imaging collaboration (isic),'' \emph{arXiv preprint arXiv:1902.03368}, 2019.

\bibitem{bernard2018deep}
O.~Bernard, A.~Lalande, C.~Zotti, F.~Cervenansky, X.~Yang, P.-A. Heng, I.~Cetin, K.~Lekadir, O.~Camara, M.~A.~G. Ballester \emph{et~al.}, ``Deep learning techniques for automatic mri cardiac multi-structures segmentation and diagnosis: is the problem solved?'' \emph{IEEE TMI}, vol.~37, no.~11, pp. 2514--2525, 2018.

\bibitem{amgad2019structured}
M.~Amgad, H.~Elfandy, H.~Hussein, L.~A. Atteya, M.~A. Elsebaie, L.~S. Abo~Elnasr, R.~A. Sakr, H.~S. Salem, A.~F. Ismail, A.~M. Saad \emph{et~al.}, ``Structured crowdsourcing enables convolutional segmentation of histology images,'' \emph{Bioinformatics}, vol.~35, no.~18, pp. 3461--3467, 2019.

\bibitem{ronneberger2015u}
O.~Ronneberger, P.~Fischer, and T.~Brox, ``U-net: Convolutional networks for biomedical image segmentation,'' in \emph{Proc. MICCAI}.\hskip 1em plus 0.5em minus 0.4em\relax Springer, 2015, pp. 234--241.

\bibitem{milletari2016v}
F.~Milletari, N.~Navab, and S.-A. Ahmadi, ``V-net: Fully convolutional neural networks for volumetric medical image segmentation,'' in \emph{Proc. 3DV}, 2016, pp. 565--571.

\bibitem{miao2024cross}
J.~Miao, C.~Chen, K.~Zhang, J.~Chuai, Q.~Li, and P.-A. Heng, ``Cross prompting consistency with segment anything model for semi-supervised medical image segmentation,'' in \emph{Proc. MICCAI}.\hskip 1em plus 0.5em minus 0.4em\relax Springer, 2024, pp. 167--177.

\bibitem{ma2024segment}
J.~Ma, Y.~He, F.~Li, L.~Han, C.~You, and B.~Wang, ``Segment anything in medical images,'' \emph{Nature Communications}, vol.~15, no.~1, p. 654, 2024.

\end{thebibliography}
}

\end{document}